\UseRawInputEncoding 
\documentclass[10pt,journal,compsoc]{IEEEtran}

\ifCLASSOPTIONcompsoc
  \usepackage[nocompress]{cite}
\else
  \usepackage{cite}
\fi

\ifCLASSINFOpdf
\else
\fi
\usepackage{amsmath}
\usepackage{ntheorem}

\usepackage{amsfonts}
\usepackage{mathrsfs}
\usepackage{amssymb}
\usepackage{lettrine}
\usepackage{algorithm}
\usepackage{algorithmic}
\usepackage{graphicx}
\usepackage{array}
\usepackage{makecell}
\usepackage{cite}
\usepackage{color}
\usepackage{booktabs}
\usepackage{url}
\usepackage{multirow}
\usepackage{footmisc}
\usepackage{bbding}
\usepackage[subnum]{cases}
\usepackage[colorlinks,
            linkcolor=blue,
            anchorcolor=blue,
            citecolor=blue,
            ]{hyperref}

\theorembodyfont{\upshape}
\theoremseparator{:}
\newtheorem{theorem}{Theorem}
\newtheorem{proposition}{Proposition}
\newtheorem{lemma}{Lemma}

\newtheorem{assumption}{Assumption}

\newtheorem{corollary}{Corollary}

\begin{document}

\title{Lifting the Veil: Unlocking the Power of Depth in Q-learning}

\author{Shao-Bo Lin, Tao Li, Shaojie Tang, Yao Wang, Ding-Xuan Zhou
\thanks{Shao-Bo Lin, Tao Li and Yao Wang are with the Center for Intelligent Decision-making and Machine Learning, School of Management, Xi'an Jiaotong University, Xi'an, China. (email: sblin1983@gmail.com; littt1024@gmail.com; yao.s.wang@gmail.com).}
\thanks{Shaojie Tang is with the Naveen Jindal School of Management, The University of Texas at Dallas, Richardson, Texas, USA. (email: Shaojie.Tang@utdallas.edu).}
\thanks{Ding-Xuan Zhou is with School of Mathematics and Statistics, University of Sydney, Sydney, Australia. (email: dingxuan.zhou@sydney.edu.au.)}
\thanks{\emph{(Corresponding author: Yao~Wang)}}
}


\IEEEtitleabstractindextext{%
\begin{abstract}
With the help of massive data and rich computational resources, deep Q-learning has been widely used in   operations research and management science and
has contributed to great success in numerous applications, including recommender systems, supply chains, games, and robotic manipulation. However, the success of deep Q-learning lacks solid theoretical verification and interpretability.  The aim of this paper is to theoretically verify the power of depth in deep Q-learning.  Within the framework of statistical learning theory, we rigorously prove that deep Q-learning outperforms its traditional version by demonstrating its good generalization error bound.  Our results reveal that the main reason for the success of deep Q-learning is the excellent performance of deep neural networks (deep nets) in capturing the special properties of rewards namely,  spatial sparseness and piecewise constancy, rather than their large capacities.  In this paper, we make fundamental contributions to the field of reinforcement learning by answering to the following three questions: Why does deep Q-learning perform so well?  When does deep Q-learning perform better than traditional Q-learning? How many samples are required to achieve a specific prediction accuracy for deep Q-learning? Our theoretical assertions are verified by applying deep Q-learning in the well-known beer game in supply chain management and a simulated recommender system.
\end{abstract}
\begin{IEEEkeywords}
deep Q-learning, deep nets, reinforcement learning, generalization error, beer game, recommender system.
\end{IEEEkeywords}}

\maketitle

\IEEEdisplaynontitleabstractindextext

\IEEEpeerreviewmaketitle

\section{Introduction}
Intelligent system operations \cite{schultz2018deep}, supply chain management \cite{giannoccaro2002inventory}, production recommendation \cite{zheng2018drn}, human resource management \cite{gallego1994optimal}, games \cite{silver2016mastering}, robotic manipulation \cite{kalashnikov2018qt}, pricing \cite{gallego1994optimal}, and many other applications,   often deal with data consisting of states, actions and rewards. Developing suitable policies based on these data to improve the quality of decision-making is an important research topic in operations research and management science. For example, in product recommendations \cite{zheng2018drn}, the state
refers to a  user's current preference, the action is the recommendation of a product to the user, and the reward concerns his/her feedback on the recommendation. The goal is to find a policy that efficiently recommends products that can maximize the cumulative rewards and thus increase revenue. To address these sequential decision-making problems,  traditional studies \cite{winston2004operations} favour {\it model-driven} approaches; many models are proposed based on human ingenuity and prior knowledge and data are utilized to select a suitable model from these candidates. Such approaches benefit from theoretical analysis and interpretability  \cite{mascis2002job,rusmevichientong2006nonparametric}, which are crucial for convincing the decision-makers and persuading consumers. However, these approaches have weak predictive accuracy and are usually labor intensive, frequently requiring people to change the model, as long as some outliers are discovered.
\begin{figure}[!t]
\centering
\includegraphics*[scale=0.3]{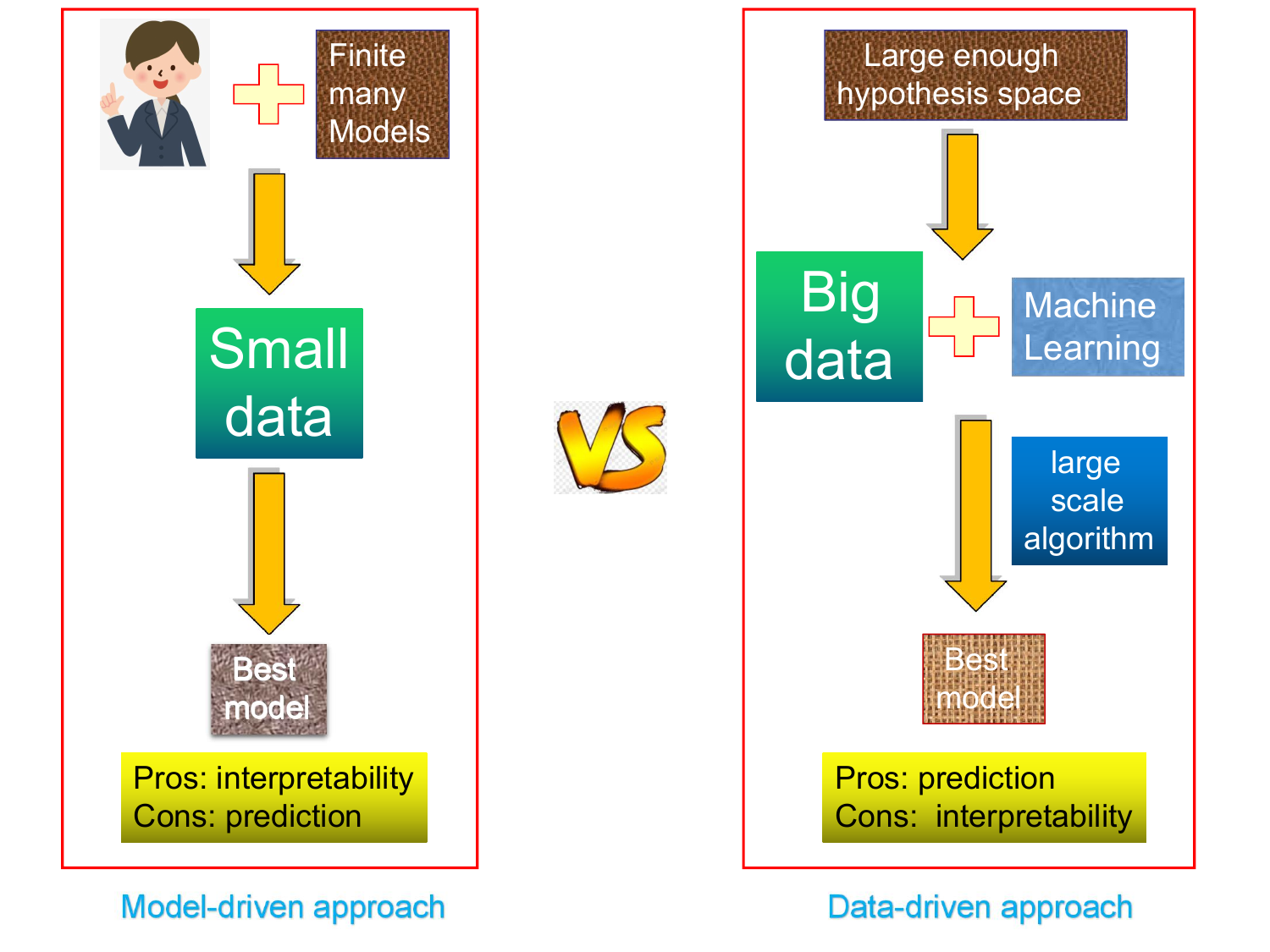}
\hfill \caption{Philosophy behind {\it model-driven} and {\it data-driven}  approaches}
\label{Fig:driven}
\end{figure}
Meanwhile, the rapid development of data mining in recent years has led to an explosion in the volume and variety of data.
These massive data certainly bring opportunities to discover subtle population patterns, but significantly impact labor-intensive {model-driven} approaches \cite{zhou2014big}.  {\it Data-driven} approaches, on the other hand, focus on utilizing machine learning methods to explore the patterns suggested by the data. They have attracted growing interest recently in operations research and management science \cite{russo2018learning,kraus2020deep,yoganarasimhan2020search,oroojlooyjadid2022deep}.  The basic idea of a data-driven approach is to first adopt a large hypothesis space, with the intuition that the space is rich enough to contain the optimal model, and then apply a specific machine learning algorithm to the massive data to tune a high-quality model from the hypothesis space.  These approaches greatly reduce the importance of human ingenuity and improve the prediction accuracy. Figure \ref{Fig:driven} differentiates the two aforementioned approaches.

However, although the {data-driven} approach has been widely applied and achieved great success in numerous applications \cite{bengio2013representation,silver2016mastering,zheng2018drn,oroojlooyjadid2022deep}, solid theoretical verification and interpretability are needed to support its excellent performance, without which decision makers' enthusiasm may be dampened and they turn to the {model-driven} approach, neglecting the outstanding performance of the {data-driven} approach in practice. Under these circumstances, a corresponding theory that explains the running mechanisms and verifies the feasibility of the {data-driven} approach should be developed urgently, which is the purpose of this paper.

\subsection{Problem formulation}

\begin{figure}[!t]
\centering
\includegraphics*[scale=0.3]{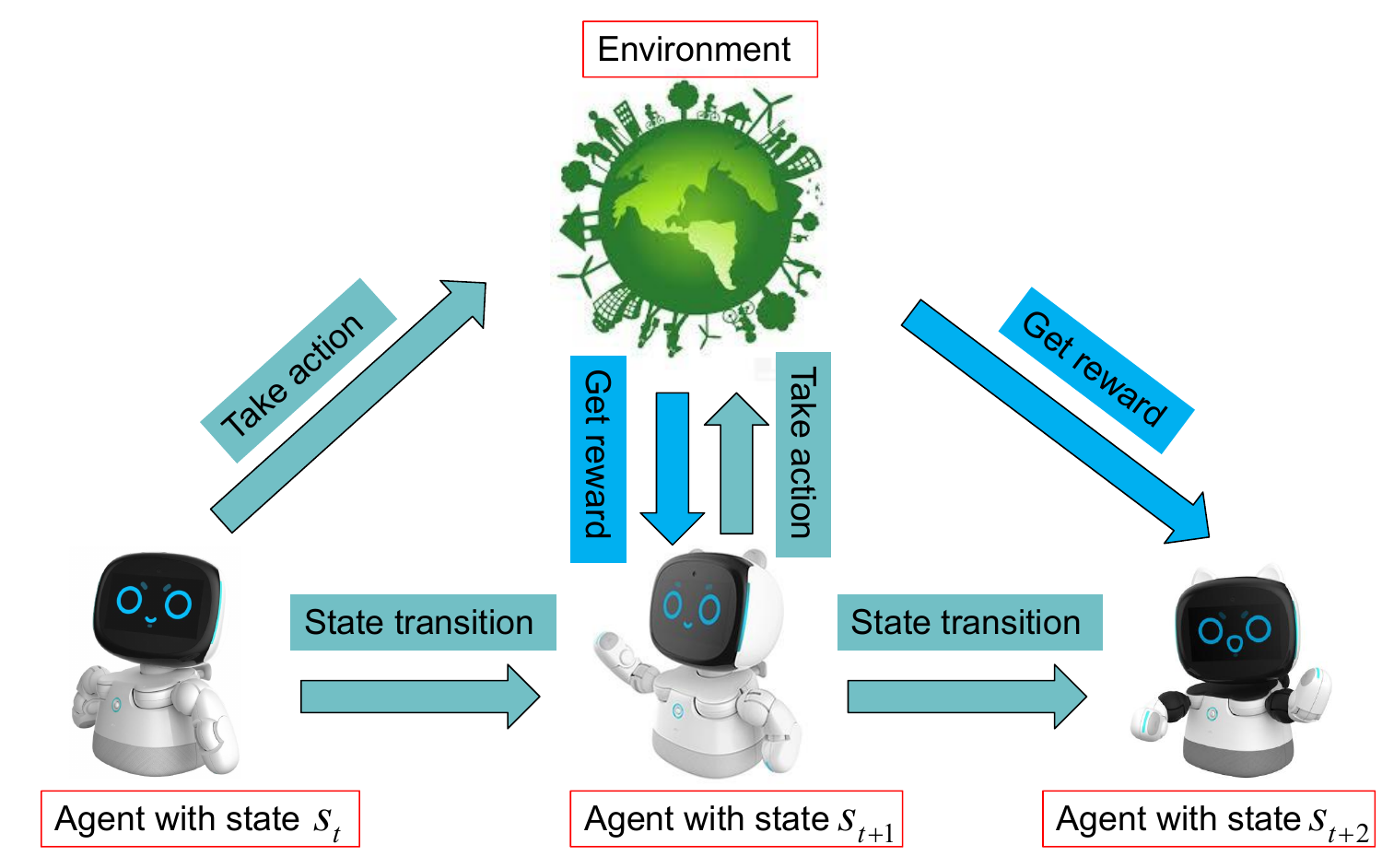}
\hfill \caption{ States, actions and rewards in RL}
\label{Fig:RL}
\end{figure}

Reinforcement learning (RL) \cite{sutton2018reinforcement} is a promising data-driven approach to tackle sequential decision-making problems with data consisting of states, actions, and rewards. As shown in Figure \ref{Fig:RL}, RL aims to develop a  sequence
of actions to maximize the total cumulative reward. It has been successfully used in human resource operations  \cite{popescu2007dynamic}, recommendations \cite{zheng2018drn}, games \cite{silver2016mastering}, supply chains \cite{giannoccaro2002inventory}, among others. The review \cite{gosavi2009reinforcement} provides additional details on RL.

The Q-learning \cite{watkins1992q} algorithm
is widely used  to produce a  sequence  of  high-quality actions in RL  and
is regarded to be one of the most important breakthroughs
in RL \cite[Sec. 6.5]{sutton2018reinforcement}.   The core of Q-learning is to learn a sequence of Q-functions that are approximations of the optimal values. Q-learning is, by definition,  a temporal-difference algorithm that incorporates four components: hypothesis space selection, optimization strategy designation, Q-functions training, and policy search. The hypothesis space component is devoted to selecting a suitable class of functions that regulate some a-priori knowledge of Q-functions. The optimization component entails the mathematical formulation of a series of optimization problems concerning   Q-functions based on the given data and selected hypothesis space. The Q-functions component aims to successively determine  Q-functions at different times by solving the formulated optimization problems.  The policy component determines the policy by maximizing the obtained   Q-functions.

As a starting point of Q-learning, the hypothesis space plays a crucial role in Q-learning because it not only regulates the format and property of the Q-function to be learned but also determines the difficulty of solving the corresponding optimization problems. Since optimal Q-functions are unknown, the selection of hypothesis spaces is an eternal problem in Q-learning \cite{sugiyama2015statistical}. In particular,
 large hypothesis spaces are beneficial for representing any Q-functions but inevitably require large computations and are frequently unstable to noise. Conversely, excessively small hypothesis spaces lack expressive powers and usually exclude the optimal Q-functions, causing Q-learning to underperform, even in fitting the given data.
Due to the limitation of computation and memory requirements,   linear spaces spanned by certain basis functions are often selected as hypothesis spaces in traditional Q-learning \cite{sutton2018reinforcement}, resulting in several bottlenecks in practice. For example, in robotic grasping  \cite{peters2008reinforcement}, Q-learning with linear hypothesis spaces is only  applicable to training individual skills, such as hitting a ball, throwing, or opening a door.

With the development of rich computational resources, such as the computational power of modern graphics processing units (GPUs), adopting large hypothesis spaces in Q-learning has become practically feasible.   Deep Q-learning, which adopts deep neural networks (or {\it deep nets}) to build hypothesis spaces, has been used in numerous applications. For instance,  AlphaGo \cite{silver2016mastering}  beats a professional human ``go" player in a complete game of Go without handicap; deep robotic grasping \cite{kalashnikov2018qt} achieved a 96\% grasp success rate on unseen objects, and certain bottlenecks in traditional Q-learning have been overcome \cite{peters2008reinforcement}.

Despite its great success in practice, deep Q-learning lacks the theoretical foundations to provide the guarantees required by many applications.  Consequently, the ability of deep Q-learning to outperform existing schemes is unclear, and users hesitate to
adopt deep Q-learning in safety-critical learning tasks, such as clinical diagnosis and financial investment.
The following  three crucial questions  should be answered to increase the confidence of decision-makers:

$\diamond$
{\bf Question 1.} Why does deep Q-learning perform so well?

$\diamond$
{\bf Question 2.}  When does deep Q-learning perform better than traditional Q-learning?

$\diamond$
{\bf Question 3.}  How many samples are required to achieve a specific prediction accuracy for deep Q-learning?

\subsection{Our Contributions}

An intuitive explanation for the success of deep Q-learning is the large capacity of deep nets \cite{bengio2013representation,kraus2020deep}, which improves the expressive power of traditional linear hypothesis spaces. In this paper, we demonstrate that this large capacity is not the determining factor, since such a feature makes the learning scheme sensitive to noise and thus leads to large variances \cite{cucker2007learning}.
We rigorously prove that the success of deep nets in Q-learning is due to their excellent performance in capturing the locality, spatial sparseness, and piecewise smoothness of rewards, which is beyond the capability of shallow nets and linear models \cite{chui1994neural}. In particular, after analyzing the relationship between optimal Q-functions and rewards, we find that optimal Q-functions are generally spatially sparse and piecewise smooth. With similar capacities, deep nets outperform shallow nets and linear models by providing a considerably better approximation error. Our main contributions are as follows.

$\bullet$ {\it Oracle inequality for Q-learning:}   An oracle inequality is
 a bound on the risk of an
estimator that shows that the performance of the estimator is almost
 as good as it would be if the
decision-maker had access to an oracle that knows what the best
the model should be.  It is an important tool that determines whether the estimator in hand
is optimal. Our first contribution is the establishment of an oracle inequality for Q-learning, which shows the crucial role of hypothesis spaces. We adopt two conflicting quantities, namely, approximation error and covering numbers, to illustrate a dilemma in selecting hypothesis spaces. The optimal performance is achieved by balancing these two quantities.

$\bullet$ {\it Expressivity for deep nets without enlarging the capacity:} Generally speaking, it is difficult to select hypothesis spaces with both small approximation errors and covering numbers. However, in Q-learning,
 optimal Q-functions depend  heavily on rewards, and the  adopted  rewards
 are usually spatially sparse and piecewise smooth \cite{sutton2018reinforcement}.
Our results rigorously demonstrate the advantage of deep nets in approximating spatially sparse and piecewise-smooth functions.
The approximation capability of deep nets is substantially better than that of linear models and shallow nets that have almost the same capacities. This finding addresses Question 1 and shows that the reason why deep Q-learning outperforms traditional Q-learning is due to its innovation of selecting hypothesis spaces that possess both small approximation errors and covering numbers, provided the rewards are spatially sparse and piecewise-smooth functions. With this, we provide a basis for answering Question 2, that is, understanding when deep Q-learning outperforms traditional Q-learning.

$\bullet$ {\it Generalization error for deep Q-learning:} Combining the approximation error estimate and the established oracle inequality, we succeed in deriving a tight generalization error bound for deep Q-learning and answering Question 3. Since deep nets can capture the spatial sparseness and piecewise smoothness of rewards, the derived generalization error is smaller than that of traditional Q-learning with linear hypothesis spaces, which shows the power of depth in deep Q-learning.

$\bullet$ {\it Numerical verifications:} Motivated by \cite{oroojlooyjadid2022deep} and \cite{ie2019reinforcement}, we apply deep Q-learning to a classical supply chain management problem: the beer game, and a simulated recommender system. We numerically show that if the reward is incorrectly specified, then deep Q-learning cannot essentially improve the performance of the classical (shallow) Q-learning approach. The effectiveness and efficiency of deepening the network are based on the appropriate reward. A similar conclusion holds for the role of data size in deep Q-learning in terms that massive data can embody the advantage of deep Q-learning, whereas small data cannot do it. These interesting phenomena numerically verify our theoretical assertions that the properties of rewards, depth of neural networks, and size of data are three crucial factors that guarantee the excellent performance of deep Q-learning.

\subsection{Organization }

 The rest of this paper is organized as
follows. In Section \ref{Sec.Q}, we explain RL and Q-learning and present a novel oracle inequality for Q-learning. In Section \ref{Sec.Deep}, we introduce several important properties of optimal Q-functions and show the performance of deep nets in approximating these Q-functions.
In Section \ref{Sec.dql}, we pursue the power of depth by exhibiting a tight generalization error of deep Q-learning.
  In Section \ref{Sec.Experiment}, we conduct experiments by applying deep Q-learning to the beer game and recommender system to verify our theoretical assertions.
We draw a simple conclusion in the last section. The proofs of our results and the background of the beer game and recommender system application are presented in the supplementary material.

\section{Related work}
Over the past decade, RL has been shown to be a powerful tool for addressing sequential decision-making problems. An important reason for this is its integration with deep nets  \cite{franccois2018introduction}. Although the power of depth in deep Q-learning remains undetermined,  there are numerous studies on the generalization capability of traditional Q-learning, the power of depth in supervised learning, and the feasibility of deep Q-learning, which are related to our work.

Unlike classical works \cite{tsitsiklis1997analysis,even2003learning,yu2009convergence} describing the convergence of Q-learning for finite-horizon RL problems, the present paper focuses on the generalization capability,  which quantifies the relationship between the prediction accuracy and number of samples. The generalization capability of Q-learning, measured by the generalization error (or finite-sample bounds), has proven a stumbling block for an enormous number of research activities over the past twenty years \cite{murphy2005generalization,kearns1999finite,goldberg2012q,dalal2018finite,zou2019finite,wainwright2019variance}. Among these works, \cite{murphy2005generalization,goldberg2012q} are the most related to our work, where the generalization error of batch  Q-learning is deduced for finite-horizon sequential decision-making problems. The novelty of our work is that we
highlight the important role of hypothesis spaces in Q-learning rather than assuming the hypothesis spaces to be linear models. Furthermore, under the same conditions, our derived generalization error bound is much smaller than those of \cite{murphy2005generalization,goldberg2012q}, since several new techniques have been developed, such as a novel error decomposition and a new concentration inequality for Q-learning.

The approximation capability of neural networks is a classical research topic that back to the 1980s, when the universal approximation property of shallow nets was established by \cite{cybenko1989approximation}. A common consensus is that the power of depth depends heavily on the properties of target functions. If the target function is smooth, then
it was proved by \cite{guo2019realizing} that deep nets perform similarly to shallow nets, showing that there is no essential improvement when the approximation tools change from shallow nets to deep nets. For complicated functions,
 deep nets have been shown to be much better than shallow nets and linear models.
In particular, with a comparable number of free parameters,
Deep nets have been proven to outperform shallow nets and linear models in embodying the manifold structures of the input space \cite{shaham2018provable}, realizing the sparseness in the frequency domain \cite{schwab2019deep} and spatial domain \cite{lin2018generalization}, reflecting the rotation-invariance of the data \cite{chui2019deep}, grasping the hierarchical features of the data \cite{mhaskar2016deep}, and approximating non-smooth functions \cite{petersen2018optimal}. All these demonstrate the power of depth in supervised learning. In this paper, we aim to derive the advantage of deep nets in approximating optimal Q-functions, which are piecewise smooth and spatially sparse in numerous applications  \cite{sutton2018reinforcement}. We rigorously prove that deep nets outperform shallow nets and linear models in approximating such Q-functions and show the reasonableness and efficiency of building  hypothesis spaces using deep nets.

To the best of our knowledge, \cite{fan2020theoretical} is the first work to show the feasibility of deep Q-learning in terms of generalization. Under some composition and smoothness assumptions for optimal Q-functions and a concentration coefficient assumption regarding marginal probability, \cite{fan2020theoretical} derived a generalization error estimate of deep Q-learning and showed that deep Q-learning beats the classical version, which is, of course, a beneficial development that lays a stepping-stone toward understanding deep Q-learning. There are three main differences between our work and that of \cite{fan2020theoretical}. The first difference is the setting; we are concerned with finite-stage sequential decision-making problems, whereas \cite{fan2020theoretical} considered infinite-stage problems involving strict restrictions on the discount factor to guarantee the concentration coefficient assumption. The second difference is the adopted algorithms, although both are variants of batch Q-learning. To be detailed, since infinite-stage sequential decision-making problems are considered in  \cite{fan2020theoretical}, the policy length was a main parameter and depended heavily on the concentration coefficient assumption, which is difficult to verify in practice; this is not the case in our analysis. The last difference is the assumptions of the optimal Q-functions; our assumptions are induced from numerous deep Q-learning applications, which is beyond the scope of \cite{fan2020theoretical}.

Another recent paper \cite{oroojlooyjadid2022deep} studied the performance of deep Q-learning in inventory optimization problems. The  basic idea was to adopt a shaped variant of rewards, with which they showed that deep Q-learning
can essentially improve the performance of classical (shallow) Q-learning. Our numerical studies are motivated by \cite{oroojlooyjadid2022deep}. However, unlike \cite{oroojlooyjadid2022deep}, which showed the outperformance of shaped deep Q-learning, we numerically elucidate the roles of depth, rewards and data size. We apply a similar numerical experiment in a simulated recommender system \cite{ie2019recsim}.  
Our numerical results aim to reveal the veil of the success of deep Q-learning. More importantly, we derive a solid theoretical analysis to show why and when deep Q-learning outperforms classical (shallow) Q-learning. Our theoretical results can provide a theoretical guarantee for  \cite{oroojlooyjadid2022deep} and \cite{ie2019reinforcement}.

\section{Oracle Inequality for  Q-learning}\label{Sec.Q}
We present an oracle inequality to show the important role of hypothesis spaces in Q-learning. Throughout the paper, we use upper-case letters to denote random variables and lower-case letters to denote instances of random variables.

\subsection{RL and Q-learning}
We consider  $T$-stage sequential decision-making problems. For $t=1,\dots,T$, let $\tilde{\mathcal S}_t\subset\mathbb R^{d_{s,t}}$ and $\tilde{\mathcal A}_t\subset\mathbb
R^{d_{a,t}}$ be families of states and actions, respectively,  in stage $t$, where $d_{s,t}, d_{a,t}\in\mathbb N$ denote the dimensions of the state and action spaces at stage $t$. The data in RL are formed as $\mathcal T_T=(\mathbf s_{T+1},\mathbf a_T,\mathbf R_T)$, where $\mathbf
s_t=\{s_1,\dots,s_t\}$ with $s_t\in \tilde{\mathcal S}_t$ is the sequence of $t$ states, $\mathbf a_t=\{a_1,\dots,a_t\}$ with
$a_t\in \tilde{\mathcal A}_t$ is the sequence of $t$ actions,  and $\mathbf R_t=\{R_1,\dots,R_t\}$ with $R_t:=R_t(\mathbf s_{t+1},\mathbf a_t)$ is the sequence of $t$ rewards.
Denote by $D:=\{\mathcal
T_{T,i}\}_{i=1}^m$   the training set of size $m$. RL   aims to derive
 $T$ maps as
$$
  \pi_t:
 \tilde{\mathcal S}_1\times\cdots\times \tilde{\mathcal S}_t \times
 \tilde{\mathcal A}_1\times\cdots\times\tilde{\mathcal A}_{t-1}\rightarrow\tilde{\mathcal
 A}_t, t=1,2,\dots,T
$$
 to maximize
$\sum_{t=1}^{T}R_t(\mathbf s_{t+1},\pi_t(\mathbf s_t,\mathbf a_{t-1}))$ based on   $D$.

Under the standard statistical framework of RL   \cite{murphy2005generalization,goldberg2012q,sugiyama2015statistical},  samples in $\{\mathcal
T_{T,i}\}_{i=1}^m$ are assumed to be  drawn independently and identically  (i.i.d.) according to a definite but unknown
distribution
\begin{equation}\label{likelihood under P}
    \begin{split}
    &P=\rho_1(s_1)p_1(a_1|s_1)
    \\&\prod_{t=2}^{T}\rho_t(s_t|\mathbf
     s_{t-1},\mathbf a_{t-1})p_t(a_t|\mathbf s_t,\mathbf
     a_{t-1})\rho_{T+1}(s_{T+1}|\mathbf s_T,\mathbf a_T),
    \end{split}
\end{equation}
where $\rho_t(s_t|\mathbf s_{t-1},\mathbf a_{t-1})$ is the conditional density of $s_t$ conditioned on $\mathbf s_{t-1},\mathbf a_{t-1}$ and $p_t(a_t|\mathbf s_t,\mathbf
     a_{t-1})$ denotes the probability that action $a_t$ is
taken given the history $\{\mathbf s_t,\mathbf a_{t-1}\}$.
A policy formed by a sequence of decision rules is written as  $\pi=\{\pi_1,\dots,\pi_T\}$. We further denote
\begin{equation}\label{likelihood under Ppi}
     \begin{split}
     &P_\pi=\rho_1(s_1)1_{a_1=\pi_1(s_1)}
     \\&\prod_{t=2}^{T}\rho_t(s_t|\mathbf
     s_{t-1},\mathbf a_{t-1})1_{a_t=\pi(\mathbf s_t,\mathbf
     a_{t-1})}\rho_{T+1}(s_{T+1}|\mathbf s_T,\mathbf a_T),
     \end{split}
\end{equation}
where for a predicate $W$, $1_W$ is 1 if $W$ is true and  0
otherwise.

Define the $t$ value function (value function at time $t$) of $\pi$ by
\begin{equation*}
    \begin{split}
     V_{\pi,t}(\mathbf s_t,&\mathbf a_{t-1})=
     \\&E_\pi\left[\sum_{j=t}^TR_j(\mathbf S_{j+1},\mathbf A_j)\big|\mathbf S_{t}=\mathbf s_t,
     \mathbf A_{t-1}=\mathbf a_{t-1}\right],
     \end{split}
\end{equation*}
where $E_\pi$ is the expectation  with respect to the distribution
$P_\pi$.
If $t=1$, then this can be written as $V_{\pi}(s_1)=V_{\pi,1}(s_1)$  for brevity. We further denote the
optimal  $t$ value function as
$$
     V^*_{t}(\mathbf s_t,\mathbf a_{t-1})=\max_{\pi\in\Pi}V_{\pi,t}(\mathbf s_t,\mathbf a_{t-1}),
$$
where $\Pi$ denotes the collection of all policies.
The Bellman equation \cite{bellman1966dynamic} characterizes the
optimal policy $\pi^*$ as
\begin{equation}\label{Bellman}
    \begin{split}
    \pi_t^*(\mathbf s_t,\mathbf a_{t-1})&=
    \arg\max_{a_t}
    E[R_t(\mathbf S_{t+1},\mathbf A_t)\\&+V_{t+1}^*(\mathbf S_{t+1},\mathbf A_t)|\mathbf S_t=\mathbf
    s_t,\mathbf A_t=\mathbf a_t],
    \end{split}
\end{equation}
where $E$  is the expectation with respect to $P$.
 RL then aims to find a policy $\hat{\pi}$ to
 minimize $V^*(s_1)-V_{\hat{\pi}}(s_1)$, where $V^*(s_1):=V_1^*(s_1)$.

Batch Q-learning
  \cite{murphy2005generalization,sutton2018reinforcement},   a widely used variant of Q-learning,
divides a $T$-stage sequential  decision-making problem into $T$ least squares problems.
The optimal
time-dependent $Q$-function is defined by
\begin{equation}\label{Qfunction}
    \begin{split}
     Q_t^*(\mathbf s_t,\mathbf a_t)&=E[R_t(\mathbf S_{t+1},\mathbf A_t)\\&+V_{t+1}^*(\mathbf
     S_{t+1},\mathbf A_t)|\mathbf S_t=\mathbf s_t,\mathbf
     A_t=\mathbf a_t].
    \end{split}
\end{equation}
Since
\begin{equation*}
    \begin{split}
     &V^*_{t}(\mathbf s_t,\mathbf a_{t-1})\\=&V_{\pi^*,t}(\mathbf s_t,\mathbf a_{t-1})
     \\=&E_{\pi^*}\left[\sum_{j=t}^TR_j(\mathbf S_{j+1},\mathbf A_j)\big|\mathbf S_{t}=\mathbf s_t,
     \mathbf A_{t-1}=\mathbf a_{t-1}\right],
     \end{split}
\end{equation*}
then
\begin{equation}\label{TD}
 V_t^*(\mathbf
     s_{t},\mathbf a_{t-1})=\max_{a_t}Q_t^*(\mathbf s_t,\mathbf
     a_t).
\end{equation}
We call $V_t^*(\mathbf
     s_{t},\mathbf a_{t-1})-Q_t^*(\mathbf s_t,\mathbf
     a_t)$  the optimal advantage
     temporal-difference at time $t$. Furthermore, according to
     \cite[Lemma 1]{murphy2005generalization} (see also \cite[Chap. 5]{kakade2003sample}),
\begin{equation}\label{Generalization}
    \begin{split}
     V^*(s_1)-&V_{\pi}(s_1)=\\&E_{\pi}\left[\sum_{t=1}^TV_t^*(\mathbf
     S_t,\mathbf A_{t-1})-Q_t^*(\mathbf S_t,\mathbf
     A_t)\big|S_1=s_1\right]
    \end{split}
\end{equation}
holds for an arbitrary $\pi$.
This equality shows that the quality of a policy $\pi$ depends on
the temporal difference, and a good estimate of optimal Q-function helps reduce the generalization error of RL.
Under these circumstances, the estimation of $Q^*_t$ for $t=1,\dots,T$ lies at the core of batch Q-learning.
With $ Q_{T+1}^*=0$, it follows from (\ref{Qfunction}) and (\ref{TD}) that for $t=T,T-1,\dots,1$,
\begin{equation}\label{conditonal expectation}
    \begin{split}
        Q_t^*(\mathbf s_t&,\mathbf a_t)
     = E[R_t(\mathbf S_{t+1},\mathbf A_t)
     \\&+\max_{a_{t+1}}Q_{t+1}^*(\mathbf S_{t+1},\mathbf
     A_{t},a_{t+1})\big|\mathbf S_t=\mathbf s_t,\mathbf A_t=\mathbf a_t ]
    \end{split}
\end{equation}

This implies the following proposition.
\begin{proposition}\label{Prop:Regression-fun}
Let $L_t^2$ be the space of square-integrable functions
with respect to the   distribution
\begin{equation}\label{likelihood}
     P_t=\rho_1(s_1)p_1(a_1|s_1)\prod_{j=2}^{t}\rho_j(s_j|\mathbf
     s_{j-1},\mathbf a_{j-1})p_j(a_j|\mathbf s_j,\mathbf
     a_{j-1}).
\end{equation}
Then, for $t=T,T-1,\dots,1$,we have
\begin{equation}\label{population}
    \begin{split}
      Q_t^*&(\mathbf s_t,\mathbf a_t)={\arg\min}_{Q_t\in L_t^2}E[(R_t(\mathbf S_{t+1},\mathbf A_t)\\&+\max_{a_{t+1}}Q_{t+1}^*(\mathbf S_{t+1},\mathbf
     A_{t},a_{t+1})-Q_t(\mathbf S_t,\mathbf A_t))^2].
    \end{split}
\end{equation}
\end{proposition}

According to Proposition \ref{Prop:Regression-fun},   optimal $Q$-functions  can be obtained by solving
$T$ least squares problems via  the backward recursion, that is,
$t=T,T-1,\dots,1$.
Empirically, by setting $\hat{Q}_{T+1}=0$,  we can compute
$Q$-functions ($\hat{Q}_t$ for $t=T,T-1,\dots,1$)   by  solving the   following
  $T$ least squares problems
\begin{equation}\label{Q learning algorithm for Q}
    \begin{split}
     \hat{Q}_t(\mathbf s_t&,\mathbf a_t) =\arg\min_{Q_t\in\tilde{\mathcal Q}_t}\mathbb E_m[(R_t(\mathbf S_{t+1},\mathbf A_t)
     \\&+\max_{a_{t+1}}\hat{Q}_{t+1}(\mathbf S_{t+1},\mathbf
     A_{t},a_{t+1})-Q_t(\mathbf S_t,\mathbf A_t))^2],
     \end{split}
\end{equation}
where $\mathbb E_m$ is the empirical expectation   and
$\tilde{\mathcal Q}_t$ is a parameterized hypothesis space.  With this, we
obtain a sequence of estimators $\{\hat{Q}_T,\dots,\hat{Q}_1\}$ and
define the corresponding policy by
\begin{equation}\label{Q learning algorithm for policy}
   \hat{\pi}_t(\mathbf s_t,\mathbf
   a_{t-1})={\arg\max}_{a_t\in\tilde{\mathcal A}_t}\hat{Q}_t(\mathbf s_t,\mathbf
   a_{t-1},a_t), t=1,\dots,T.
\end{equation}

\subsection{Oracle inequality  for Q-learning}
In this subsection, we aim to derive our novel oracle inequality for  Q-learning.
 We first introduce two mild assumptions. 
\begin{assumption}\label{Assumption:case-1}
Let $\mu\geq 1$ be a constant. Then,
\begin{equation}\label{assumption-1}
   p_t(a|\mathbf s_t,\mathbf a_{t-1})\geq \mu^{-1},\qquad\forall  a\in\tilde{\mathcal A}_t.
\end{equation}
\end{assumption}

Assumption \ref{Assumption:case-1} is a standard assumption made in \cite{murphy2005generalization,goldberg2012q} and it states
that every action in $\tilde{\mathcal A}_t$ has a positive conditional probability of being chosen at each time $t$. It contains at least two widely used settings. One is that  $\tilde{\mathcal A}_t$ only contains finitely many actions, which is the case in go games \cite{silver2016mastering}, blackjack \cite{sutton2018reinforcement},  robotic grasping \cite{kalashnikov2018qt} and beer game \cite{oroojlooyjadid2022deep}. The other is that $\tilde{\mathcal A}_t$ is an infinite set, but only finite actions in $\tilde{\mathcal A}_t$ are active when in the case of $\{\mathbf s_t,\mathbf a_{t-1}\}$. For example, in a recommender system, if the feedback for a client's age is around three, then the following candidate action is to recommend children's products. Hence, Assumption \ref{Assumption:case-1} is a mild condition that can be easily satisfied for numerous applications.
Since rewards are given before the learning process and are always assumed to be finite, we present the following mild assumption.
\begin{assumption}\label{Assumption:bounded-Reward}
 There exists a $U>0$ such that  $\|R_t\|_{L^\infty}\leq U$ for any $t=1,\dots,T$.
\end{assumption}

According to (\ref{conditonal expectation}) and    $Q_{T+1}^*=0$, Assumption \ref{Assumption:bounded-Reward} implies that
$\|Q^*_t\|_{L^\infty}\leq 2U$ for all $t=1,2,\dots,T$. Therefore, it is natural to search for estimators uniformly bounded by $2U$. To describe the role of hypothesis space, we also introduce the empirical covering number \cite{rusmevichientong2006nonparametric,gyorfi2006distribution} to quantify its capacity.  For a set of functions $\mathcal G$ defined on $\mathcal X\subseteq \mathbb R^d$ with  $d\in\mathbb N$, denote by  $\mathcal N_1(\epsilon,\mathcal G,x_1^m)$  with   $x_1^m=(x_1,\dots,x_m)\in\mathcal X^m$ the $\ell^1$ empirical covering number \cite[Def.9.3]{gyorfi2006distribution} of $\mathcal G$, which is the number of elements in a least $\varepsilon$-net of $\mathcal G$ with respect to $\|\cdot\|_{\ell^1}$. Furthermore, let $\mathcal N_1(\epsilon,\mathcal G):=\max_{x_1^m\in\mathcal X^m}\mathcal N_1(\epsilon,\mathcal G,x_1^m)$. We then obtain
  the following oracle inequality for batch Q-learning.

\begin{theorem}\label{Theorem:oracle}
Let  $\beta_t>0$, $\tilde{\mathcal Q}_t$, $t=1,\dots,T$ be  sets of functions uniformly bounded by $2U$ and $\tilde{\mathcal Q}_{T+1}=\{0\}$. If Assumptions \ref{Assumption:case-1} and \ref{Assumption:bounded-Reward} hold  and  $\hat{Q}_t$ is defined by (\ref{Q learning algorithm for Q}), then
\begin{equation}\label{generalization-error-general}
    \begin{split}
    &E[V^*(S_1)-V_{ {\pi}}(S_1)]
        \leq
       \\&C\sum_{t=1}^T  \sum_{j=t}^T  (3\mu)^{j-t}( \min_{h_j\in\tilde{\mathcal Q}_j}E[(h_j-Q_j^*)^2]+\beta_j
       + \frac{1}{m}
         \\& +\frac{1}{m }\exp\left(-{C'\beta_j m}\right) ({\cal N}_1(C'\beta_j,  \tilde{{\cal Q}}_{j})+{\cal N}_1( {C}'\beta_j, \tilde{{\cal Q}}_{j+1})))^\frac12,
    \end{split}
\end{equation}
where $C$ and $C'$ are constants depending only on $U$.
\end{theorem}

Theorem \ref{Theorem:oracle} presents a bias-variance trade-off in selecting the hypothesis space $\tilde{\mathcal Q}_t$. If $\tilde{\mathcal Q}_t$ is large, then the approximation error $ \min_{h_j\in\tilde{\mathcal Q}_j}E[(h_j-Q_j^*)^2]$ is small but the capacity ${\cal N}_1(C'\beta_j,  \tilde{{\cal Q}}_{j})$ will be large, leading to bad generalization. Conversely, if $\tilde{\mathcal Q}_t$ is small, then  ${\cal N}_1(C'\beta_j,  \tilde{{\cal Q}}_{j})$ is small but its approximation performance is not so good, which will also result in bad generalization. A suitable hypothesis space should be selected to
balance the bias and variance in each stage,   thereby achieving the best learning performance. The bias-variance balance is determined by the a-priori information of $Q^*_t$,  without which it is impossible to derive a satisfactory generalization error  \cite[Chap. 3]{gyorfi2006distribution}.

Estimation of the capacity of a hypothesis space is a classical topic in statistical learning theory \cite{gyorfi2006distribution,cucker2007learning,steinwart2008support}. In particular, it is discussed in \cite[Chap. 9]{gyorfi2006distribution}   for linear models of dimension $k$, $\mathcal G^{\diamond}_{k}$, and \cite[Chap. 16]{gyorfi2006distribution} for shallow nets $G_k^*$ with $k$ tunable parameters and some specified activation function for which
\begin{equation}\label{Covering-number-linear}
     \log \mathcal N_1(\varepsilon, \mathcal G_{k,M})\leq C_1k\log \frac{M}{\varepsilon},\qquad \forall \varepsilon>0,
\end{equation}
where $\mathcal G_{k,M}=\{f\in\mathcal G_K,\|f\|_\infty\leq M\}$, $\mathcal G_k$ is either $\mathcal G_k^*$ or $\mathcal G_k^\diamond$, $M>0$ and $C_1$ is a constant depending only on $M$.
  The following corollary then follows from Theorem \ref{Theorem:oracle} with $\beta_t=\frac{2C_1\max\{k_t,k_{t+1}\}\log(2C'Um)}{C'm}$.
\begin{corollary}\label{Corollary:generalization-for-Q}
Let  $k_t\in\mathbb N$, $k_{T+1}=0$, $\tilde{\mathcal Q}_{T+1}=\{0\}$ and $\tilde{\mathcal Q}_t$, $t=1,\dots,T$ be  sets of functions satisfying (\ref{Covering-number-linear}) with $k=k_t$ and $M=2  U$.  If Assumptions  \ref{Assumption:case-1} and \ref{Assumption:bounded-Reward} hold  and $\hat{Q}_t$ is defined by (\ref{Q learning algorithm for Q}), then
\begin{equation*}
    \begin{split}
    E[&V^*(S_1)-V_{ {\pi}}(S_1)]
         \leq
       C^{''}\sum_{t=1}^T  \sum_{j=t}^T  (3\mu)^{j-t}
       \\&\left( \min_{h_j\in\tilde{\mathcal Q}_j}E[(h_t-Q_t^*)^2]
       + \frac{\max\{k_t,k_{t+1}\}\log (2m)}{m}
          \right)^\frac12,
    \end{split}
\end{equation*}
where $C^{''}$ is a constant  depending only on $U$.
\end{corollary}

The oracle inequality for batch Q-learning was initially deduced by \cite{murphy2005generalization} under the same setting as ours. However, there are three crucial differences between our results and those of \cite[Theorem 1]{murphy2005generalization}. First, we do not assume that $Q_t^*\in\tilde{\mathcal Q}_t$ and utilize the approximation error to measure the expressive power of $\tilde{\mathcal Q}_t$. Since optimal Q-functions are rarely known in practice, the assumption of $Q_t^*\in\tilde{\mathcal Q}_t$ requires an extremely large hypothesis space, which leads to a large generalization error.
Then,    the derived generalization error bound in Theorem \ref{Theorem:oracle} or Corollary \ref{Corollary:generalization-for-Q} is essentially better than that of \cite[Theorem 1]{murphy2005generalization} under the same conditions. In particular, if  $\tilde{\mathcal Q}_t$ is a linear space and
$Q_t^*\in\tilde{\mathcal Q}_t$, our derived  generalization error in Corollary \ref{Corollary:generalization-for-Q}, is of order $\mathcal O(m^{-1/2})$, whereas that in \cite[Theorem 1]{murphy2005generalization} is of order $\mathcal O(m^{-1/4})$.
Finally,   we take the covering number to measure the generalization error without presenting any restrictions on it, which is totally different from \cite[Theorem 1]{murphy2005generalization} that conducted the analysis under capacity restrictions like \eqref{Covering-number-linear}. This makes our analysis applicable to numerous hypothesis spaces, such as reproducing kernel Hilbert space \cite{lin2018distributed},   shallow nets \cite[Chap.16]{gyorfi2006distribution} and deep nets \cite{guo2019realizing}.


\section{Deep Nets in Approximating Optimal Q-functions}\label{Sec.Deep}

{ In this section, we first demonstrate some good properties of optimal Q-functions of many real-world applications and then analyze the power of depth in approximating Q-functions via deep nets.}

\subsection{A priori information for optimal Q-functions}

We provide some excellent applications of deep Q-learning and present the a-priori information for optimal Q-functions.
Q-learning has been proven to gain considerably profitable (long-term) rewards \cite{shani2005mdp} in a recommender system, where the state is defined as the browsing history of a user, and the action is to recommend (one or more item) items to the user and the reward is the user's feedback, including
skipping, clicking, or ordering of these items. This implies that the reward function is piecewise constant and spatially sparse with respect to the state and action. Traditional Q-learning adopts linear models to formulate the hypothesis space,
which cannot capture piecewise constancy or spatial sparseness \cite{lin2018generalization}. This makes it difficult to design an efficient policy to recommend multiple products simultaneously \cite{zheng2018drn}. Deep Q-learning succeeds in overcoming this bottleneck of traditional Q-learning \cite{zheng2018drn} by recommending numerous high-quality products simultaneously.

Q-learning also provides a promising avenue for manipulating robotic interaction. Traditional Q-learning is only applicable to individual skills, such as hitting a ball, throwing, or opening a door \cite{peters2008reinforcement}. Consequently, \cite{kalashnikov2018qt} developed a deep Q-learning approach to robotic grasping that achieved a 96\% grasp success rate on unseen objects.  In their approach, the state includes the robot's current camera observation, and an RGB image with a certain resolution  (e.g., 472$\times$472). The action consists of a vector indicating the desired change in the gripper position, such as the open and close commands.  The reward is 1 at the end of the episode if the gripper contains an object and is above a certain height, and 0 otherwise, showing the piecewise constant and spatially sparse property of the reward functions.

Q-learning has numerous other applications, where the reward functions are set to be piecewise smooth (or piecewise constant) and spatially sparse. We refer the readers to a detailed beer game example in Section 4 of the supplementary material or some interesting examples in \cite{sutton2018reinforcement} shown in Table \ref{Table:Example}.
All these shows that piecewise smoothness (or piecewise constant) and spatial sparseness are two vital properties of reward functions.
Under this circumstance, it follows from (\ref{conditonal expectation}) and  $ Q_{T+1}^*=0$  that optimal Q-functions are also piecewise smooth (or piecewise constant) and spatially sparse, as shown in Table \ref{Table:Example}.

 \begin{table*}[!]\label{Table:Example}
 \fontsize{8.5}{8.5}\selectfont
  \caption{Propositions  for rewards   and optimal $Q$-functions for games \cite{sutton2018reinforcement}}
  \begin{center}
\begin{tabular}{|c|c|c|c|}
  \Xhline{2\arrayrulewidth}
  Games & Rewards & $Q^*$ &  References\cr
  \Xhline{2\arrayrulewidth}
  Pick-and-Place Robot & Piecewise constant & Piecewise constant & Example 3.2\cr
  Recycling Robot &Piecewise constant & Piecewise constant & Example 3.3 \cr
  Pole-Balancing& Piecewise constant & Piecewise constant & Example 3.4 \cr
  Gridworld & Piecewise constant & Piecewise constant & Example 3.5 \cr
  Golf & Piecewise constant & Piecewise constant & Example 3.6 \cr
  Gambler Problem & Piecewise constant & Piecewise constant & Example 4.3 \cr
  Blackjack & Piecewise constant & Piecewise constant & Example 5.1 \cr
  Driving Home & Piecewise smooth & Piecewise smooth &  Example 6.1 \cr
  Cliff Walking & Piecewise constant & Piecewise constant & Example 6.6 \cr
  Dyna Maze &Piecewise constant & Piecewise constant & Example 8.1 \cr
  Racetrack & Piecewise constant & Piecewise constant & Example 8.6 \cr
  Mountain Car & Piecewise constant & Piecewise constant & Example 10.1 \cr
   Access-control Queuing & Piecewise constant & Piecewise constant & Example 10.2 \cr
   AlphaGo & Piecewise constant & Piecewise constant & Sec. 16.6.1 \cr
   \Xhline{2\arrayrulewidth}
\end{tabular}
\end{center}
\end{table*}

In the following, we provide the mathematical definition of spatially sparse and piecewise smooth  (or piecewise constant) functions.
For $d,N\in\mathbb N$ and $\mathbb I^d:=[0,1]^d$, we partition $\mathbb
I^d$ by $N^d$ sub-cubes $\{A_j\}_{j=1}^{N^d}$ of side length
$N^{-1}$ and with centers $\{\zeta_j\}_{j=1}^{N^d}$. For $s\in\mathbb
N$ with $s\leq N^d$ and a function
$f$ defined on $\mathbb I^d$, if the support of $f$ is contained in
$\cup_{j\in \Lambda_s}A_{  j}$ for some subset $\Lambda_s$ of $\{1,\dots,N^d\}$ of cardinality $s$, we then say that $f$ is
$s$ spatially sparse in $N^d$ cubic partitions.

\begin{figure}[!t]
\centering
\includegraphics*[scale=0.30]{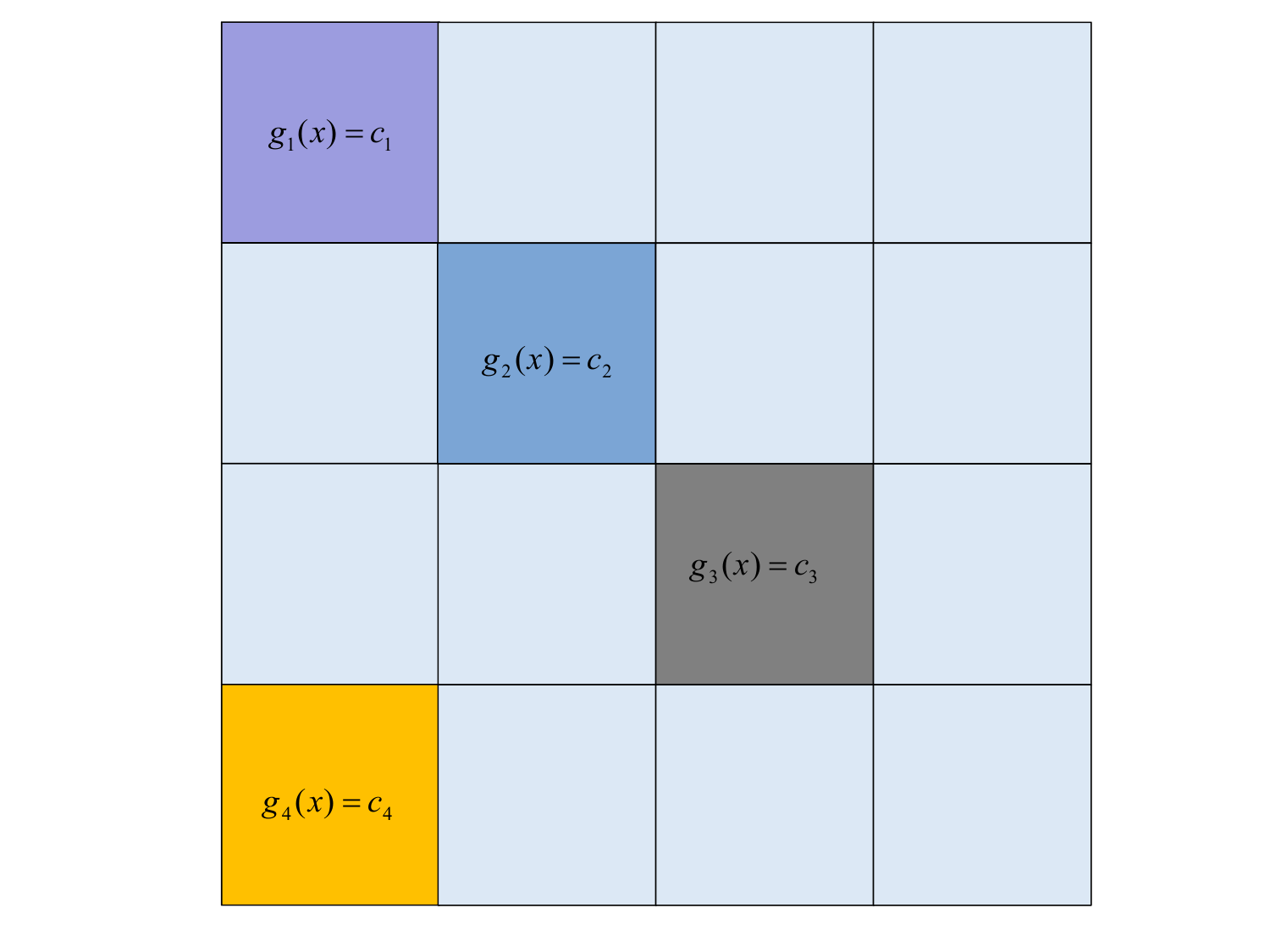}
\hfill \caption{Piecewise-constant and spatially sparse functions in $\mathcal C^{(C_0,4,16)}$}
\label{Fig:target-function}
\end{figure}

Let   $c_0>0$, $r=u+v$ with $u\in\mathbb N_0:=\{0\}\cup\mathbb N$, $0<v\leq 1$,  and $\mathbb A\subseteq\mathbb I^d$. We say that a   function $f:\mathbb
A\rightarrow\mathbb R$ is $(r,c_0)$-smooth, if $f$ is $u$-times
differentiable and for any  $\alpha = (\alpha_1, \cdots, \alpha_d) \in
{\mathbb N}^d_0$ with  $\alpha_1+\dots+\alpha_d=u$ and
          $x,x'\in\mathbb A$,  its   partial
derivative
satisfies the Lipschitz condition
\begin{equation}\label{lip}
          \left|\frac{\partial^uf}{\partial x_1^{\alpha_1}\dots\partial
          x_d^{\alpha_d}}
          (x)-\frac{\partial^uf}{\partial x_1^{\alpha_1}\dots\partial
          x_d^{\alpha_d}}
          (x')\right|\leq c_0\|x-x'\|^v,
\end{equation}
where  $\|x\|$ denotes the Euclidean norm of
  $x$. $Lip^{(r,c_0)}_{\mathbb A}$ is then written as the set of functions satisfying (\ref{lip}).
If  there exists  $g_j\in Lip^{(r,c_0)}_{A_j}$ for $j=1,\dots,N^d$ such that
\begin{equation}\label{def.piecewise-sm}
     f(x)=\sum_{j\in\Lambda_s}g_j(x)\mathcal I_{A_j}(x),
\end{equation}
 we then say that $f$ is  $s$ spatially sparse in $N^d$ cubic partitions and $(r,c_0)$-smooth, where $\mathcal I_{A_j}$ is the indicator function of $A_j$, i.e., $\mathcal I_{A_j}(x)=1$ if $x\in A_j$ and $\mathcal I_{A_j}(x)=0$ if $x\notin A_j$.
 Denote by $Lip^{(r,c_0,s,N^d)}$ the set of all such functions.
A special case of   $ f\in Lip^{(r,c_0,s,N^d)}$ is $g_j(x)=c_j$ for some $|c_j|\leq C_0$ and $C_0>0$.  In this case,
we   say that $f$ is  $s$ spatially sparse in $N^d$ cubic partitions and piecewise constant. We further denote by  $\mathcal C^{(C_0,s,N^d)}$   the set of all these functions. Figure \ref{Fig:target-function} shows a piecewise constant and spatially sparse function in $\mathcal C^{(C_0,4,16)}$.

\subsection{Capacity of deep nets}

Let $L\in\mathbb N$   and
$d_0,d_1,\dots,d_L\in\mathbb N$ with $d_0=d$. For
$\vec{h}_k=(h^{(1)},\dots,h^{(d_k)})^T\in\mathbb R^{d_k}$, define
$\vec{\sigma}(\vec{h})=(\sigma(h^{(1)}),\dots,\sigma(h^{(d_k)}))^T$, where $\sigma(t)=\max\{t,0\}$ is
  the rectified linear unit (ReLU).
Deep nets with depth $L$
  and width $d_j$ in the $j$th hidden layer can be mathematically
  represented as
\begin{equation}\label{Def:DFCN}
     h_{\{d_0,\dots,d_L\}}(x)=\vec{a}\cdot
     \vec{h}_L(x),
\end{equation}
where
\begin{equation}\label{Def:layer vector}
    \vec{h}_k(x)=\vec{\sigma}(W_k\cdot
    \vec{h}_{k-1}(x)+\vec{b}^{(k)}),\qquad k=1,2,\dots,L,
\end{equation}
$\vec{a}\in\mathbb R^{d_L}$, $\vec{b}^{(k)}\in\mathbb R^{d_k},$
$\vec{h}_{0}(x)=x,$ and $W_k=(w_{i,j}^{(k)})_{1,1}^{d_{k},d_{k-1}}$
is a $d_{k}\times
 d_{k-1}$ matrix.  The structure  of   deep nets is reflected by   the parameter
 matrices $(W_k,\vec{b}^k)$. A typical deep net is  a deep fully connected neural network (DFCN), which  corresponds to weight matrices without
any constraints.

We denote by $\mathcal H_{\{d_0,\dots,d_L\}}$ the set of all deep nets formed as (\ref{Def:DFCN}). Then,   there are totally
\begin{equation}\label{number-para}
    n_{L}:= \sum_{j=0}^{L-1} (d_jd_{j+1}+d_{j+1})+d_L
\end{equation}
tunable  parameters for   $h\in\mathcal H_{\{d_0,\dots,d_L\}}$. If $L$ is large, there are too many parameters to be tuned,   leading to extremely large capacity.
It is known that deep sparsely connected nets (DSCN), such as deep convolution neural networks \cite{zhou2020theory,zhou2020universality}, deep nets with tree structures \cite{chui2019deep} or other sparse structures \cite{petersen2018optimal}, can significantly reduce the capacity of DFCN without sacrificing its approximation capability very much.
The hypothesis spaces in this paper are DSCNs with  $n\ll n_L$ tunable parameters paved on $L$ hidden layers. Denote by $\mathcal H_{n,L}$ the set of all these deep nets with a specific structure. Thus, we define
\begin{equation}\label{hypothesis-space-proj}
     \mathcal H_{n,L,M}:=\{ h\in\mathcal H_{n,L}:\|h\|_{L^\infty}\leq M\}.
\end{equation}
The following lemma presents a covering number estimate for  $\mathcal H_{n,L,M}$.
\begin{lemma}\label{Lemma:covering-number}
  There is a constant $C^*_1$ depending only on $d$ such that
\begin{equation}\label{covering-number-deep-net}
   \log\mathcal N_1(\epsilon,\mathcal H_{n,L,M})\leq C^*_1 Ln
   \log D_{\max} \log \frac{M}\epsilon,
\end{equation}
where $D_{\max}:=\max_{0\leq j\leq L}d_j$.
\end{lemma}
Since $\|Q_t^*\|_{L^\infty}\leq 2U$, it is natural  to take $\mathcal H_{n,L,2U}$ as the hypothesis space. It should be mentioned that except for the boundedness of the neural networks,
Lemma \ref{Lemma:covering-number}  does not impose any additional restrictions on the boundedness of weights, which is different from \cite{guo2019realizing} and \cite[Chap. 14]{anthony2009neural}. If $L$ is not excessively large,   then it follows from    (\ref{covering-number-deep-net}) and (\ref{Covering-number-linear}) that the covering number of deep nets is comparable with that of shallow nets with $n$ parameters and linear models of $n$-dimension.

\subsection{Approximation capability of deep nets}

 A common consensus on deep nets' approximation  \cite{chui1994neural,mhaskar1996neural,yarotsky2017error,anthony2009neural,lin2017limitations,shaham2018provable,guo2019realizing} is that the power of depth depends on the properties of target functions. If the target function is assumed to be in $Lip^{(r,c_0)}_{\mathbb I^d}$, then
\cite{guo2019realizing} verified that deep nets perform similarly to shallow nets, showing that there are no essential improvements when the approximation tools changed from shallow nets or linear models to deep nets. However, if some additional a-priori knowledge is given, deep nets are much more effective than shallow nets and linear models, as Table \ref{Tab:ReLU_fea} shows.

\begin{table}[!h]
\fontsize{6.5}{6.5}\selectfont
\caption{Power of depth   in approximating special target functions \\(within accuracy $\varepsilon$) }\label{Tab:ReLU_fea}
\begin{center}
\begin{tabular}{|l|l|l|l|l|}
\hline Reference. & Features of target functions & Parameters & Depth\\
\hline \cite{chui2020realization} & Locality & $4d+1$ & $2$\\
\hline \cite{chui2020realization} & $k$-spatially sparse & $k(4d+1)$ & $2$  \\
\hline \cite{petersen2018optimal} &Piecewise $(r,c_0)$-smooth & $\varepsilon^{-d/r}$ & $\mathcal O(d,r)$ \\
\hline \cite{han2020depth} & $\ell_2$ radial + $(r,c_0)$-smooth & $\varepsilon^{-1/r}$ & $\mathcal O(d,r)$   \\
\hline \cite{schwab2019deep} & $k$-sparse (frequency) & $k\log(\varepsilon^{-1})$ & $\log(\varepsilon^{-1})$   \\
\hline \cite{shaham2018provable} & $d'$ dimensional manifold+smooth & $\varepsilon^{-d'/r}$ & $4$   \\
\hline
\end{tabular}
\end{center}
\end{table}

 As discussed in Sec. 3.1,   optimal Q-functions
are frequently piecewise constant (or piecewise smooth) and spatially sparse.  Studying the advantage of deep nets in approximating such functions is our main purpose, as addressed in the following theorem.

\begin{theorem}\label{Theorem:approx-piece-constant}
Let $d\geq 2$, $N\in\mathbb N$, $C_0>0$, $s\leq N^d$, $1\leq p<\infty$ and $0<\tau<0$. There exists a deep net structure with $L=2$, $\mathcal O(N^d)$ free parameters and $D_{\max}=\mathcal O(N^d)$ such that for
any $Q^*\in \mathcal C^{(C_0,s,N^d)}$,   there exists a deep net $\mathcal N_{N,s,\tau,Q^*}$ with the aforementioned structure satisfying
\begin{equation}\label{app.piece-const}
     \|Q^*-\mathcal N_{N,s,\tau,Q^*}\|_p\leq 2d C_0s \tau N^{1-d},
\end{equation}
where
$\|\cdot\|_p$ is the norm of $p$-times Lebesgue integrable function space  $L^p(\mathbb I^d)$.
\end{theorem}

\begin{figure}[!t]
\begin{minipage}[b]{0.49\linewidth}
\centering
\centerline{{\small (a) Ground truth}}
\includegraphics*[width=4cm,height=4cm]{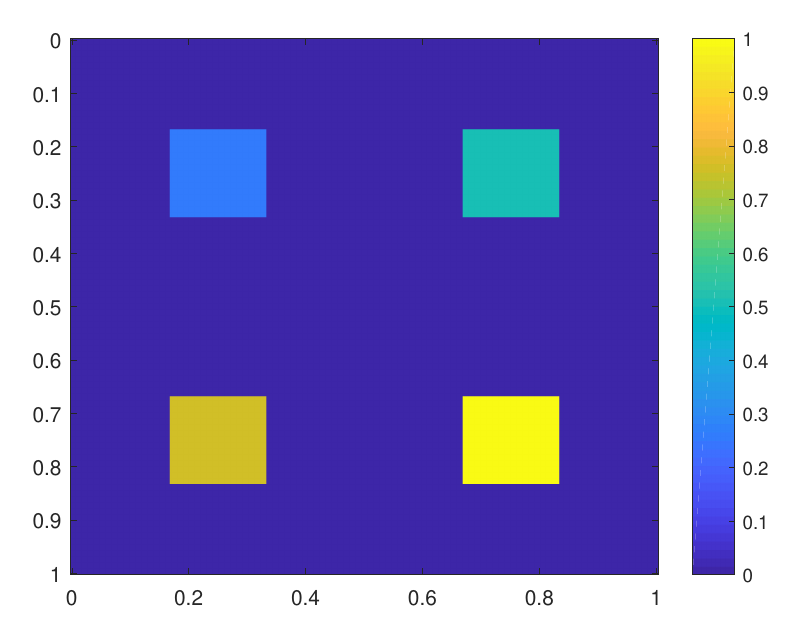}
\end{minipage}
\hfill
\begin{minipage}[b]{0.49\linewidth}
\centering
\centerline{{\small   (b)   $\tau=0.01$}}
\includegraphics*[width=4cm,height=4cm]{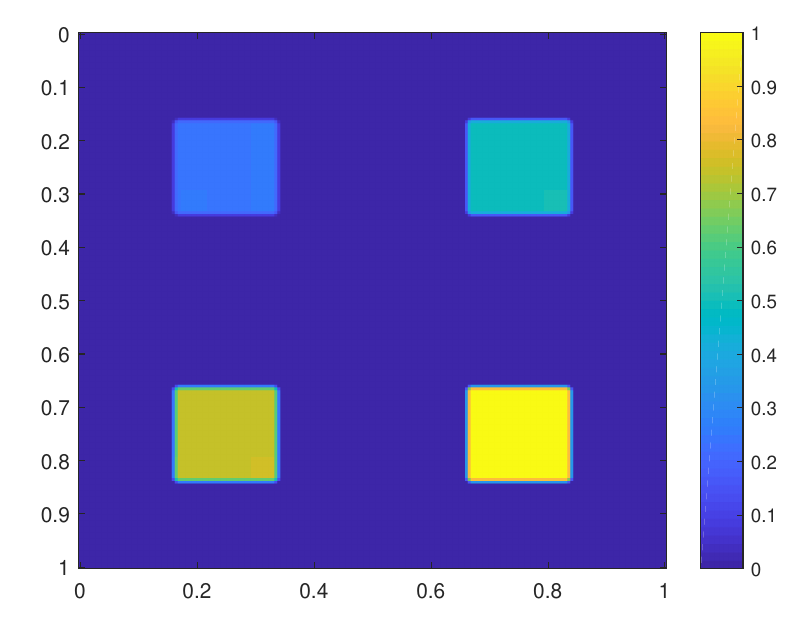}
\end{minipage}
\hfill
\caption{Performance of  deep nets constructed in Theorem \ref{Theorem:approx-piece-constant} with  $\tau=0.01$}
\label{Figure:sparse}
\end{figure}

The detailed structure of deep nets in Theorem \ref{Theorem:approx-piece-constant} is given in the proof.
Since functions in $\mathcal C^{(C_0,s,N^d)}$ are discontinuous in general,  linear models \cite{pinkus1999approximation} suffer from the  Gibbs phenomenon in the sense that the linear estimators overshoot at a jump discontinuity, and this overshoot persists as the dimension increases. Shallow nets were utilized in \cite{llanas2008constructive} to avoid
the Gibbs phenomenon when $d=1$. For $d\geq 2$, it can be found in \cite{chui1994neural,petersen2018optimal} that shallow nets are not the optimal approximation tools in the sense that they cannot achieve the optimal approximation rates. In Theorem \ref{Theorem:approx-piece-constant}, we rigorously prove that deep nets succeed in overcoming certain drawbacks of linear models and shallow nets in terms of providing perfect approximation error. In fact, we can set $\tau$ to be extremely small such that  $\|Q^*-\mathcal N_{N,s,\tau,Q^*}\|_p\leq\nu$
for arbitrarily small $\nu>0$. In a word, by adding only one hidden layer to shallow nets, we can use  $\mathcal O(N^d)$ free parameters to yield an approximant within an arbitrary accuracy, provided the target function is in $\mathcal C^{(C_0,s,N^d)}$. As stated in Lemma \ref{Lemma:covering-number}, the $\varepsilon$-covering number of deep nets is of the order $\mathcal O\left(N^d\log\frac{1}\varepsilon\right)$, which is the same as that of shallow nets with $N^d$ free parameters and linear models with $N^d$-dimension. In Figure \ref{Figure:sparse}, we provide a numerical example to show the performance of deep nets in approximating functions in $\mathcal C^{(1,4,36)}$ with   $\tau=0.01$.

In the following theorem, we pursue the power of depth in approximating piecewise smooth and spatially sparse functions.

\begin{theorem}\label{Theorem:piecewise-smooth}
Let $1\leq p<\infty$, $C_0>0$, $N,d,s\in\mathbb N$, and $r=u+v$ with $u\in\mathbb N$ and $0<v\leq 1$. There exists a
deep net  structure  with
\begin{equation}\label{number of layer}
      2(d+u)\left\lceil\frac{r+2d}{2d}\right\rceil+8(d+u)+3+\left\lceil\frac{rp+d+p+1}{2d}\right\rceil
\end{equation}
layers,  $nN^d$  free
parameters  and $D_{\max}=poly(N^dn)$ such that for any $Q^*\in Lip^{(r,c_0,s,N^d)}$,    there is a $\mathcal N_{n,s,N,Q^*}$ with the aforementioned structure  satisfying
$$
     \|Q^*-\mathcal N_{n,s,N,Q^*}\|_p
     \leq
    C_2^*n^{-r/d}sN^{-d/p},
$$
where $C_2^*$ is a constant  independent of $n,s,N$ and $poly(n)$ is a polynomial with respect to $n$.
\end{theorem}

Shallow nets were verified in    \cite{mhaskar1996neural} to be at least as good as linear models in the sense that the approximation error of shallow nets is always not larger than that of linear models. The problem is, as mentioned in \cite{chui2019deep},  that the capacity of shallow nets in \cite{mhaskar1996neural} is usually much larger than that of linear models.  This leads to the insatiability of the estimator and a large variance. Furthermore, as discussed above, both linear models and shallow nets are difficult to approximate discontinuous functions.
Theorem \ref{Theorem:piecewise-smooth}  with $s=N^d$ presents an approximation rate of deep nets when the target function is piecewise smooth, a special type of discontinuous function. It is shown that deep nets can achieve an order of $\mathcal O(n^{-r/d})$ when $p=1$, which is an optimal approximation rate \cite{petersen2018optimal} if there are $N^d$ pieces and $N^dn$ parameters.

Besides their discontinuity,   optimal Q-functions are spatially sparse, which was not considered in \cite{petersen2018optimal}. Theorem \ref{Theorem:piecewise-smooth} is devoted to approximating discontinuous and spatially sparse functions and demonstrates that deep nets outperform shallow nets by showing an additional reducing factor $sN^{-d/p}$ to reflect the sparsity in the approximation error estimate. It should be highlighted that our result is essentially different from \cite{chui2020realization}, in which the target function is continuous.

In the proofs of  Theorems \ref{Theorem:approx-piece-constant} and \ref{Theorem:piecewise-smooth}, we shall provide concrete structures of deep nets to approximate spatially sparse and piecewise smooth (or piecewise constant) functions. It should be mentioned that the structure is not unique. In fact, we can derive numerous depth-width pairs of deep nets to achieve the same approximation performance by
using the approach in \cite{han2020depth}. Furthermore, all these structures can be realized by deep convolutional neural networks via using the technique in \cite{zhou2020theory,zhou2020universality}. However, since the purpose of our study is to demonstrate the power of depth, we consider only one structure for brevity.

\section{Power of Depth in Deep Q-Learning}\label{Sec.dql}

The aim of this section is to show the power of depth in deep Q-learning.

\subsection{Learning schemes and assumptions}
The main difference between deep Q-learning and the traditional version is their hypothesis spaces. The latter uses linear models, which benefits in computation, whereas the former adopts deep nets to enhance prediction performance. To simplify our analysis, we present a Markov assumption for the distribution $P$ defined by (\ref{likelihood under P}).
\begin{assumption}\label{Assumption:Markov}
Let $Q_t^*$ be defined by (\ref{Qfunction}). We have
$$
   Q_{t}^*({\bf s}_t, {\bf a}_t)=
   Q_t^*(s_t,a_t).
$$
\end{assumption}

It should be mentioned that Assumption \ref{Assumption:Markov} is not necessary in our analysis, since our result, as shown in Theorem \ref{Theorem:oracle}, holds for an arbitrary $P$.   Without the Markov assumption,   optimal Q-functions ($Q_{t}^*$, $t=1,\dots,T$), are functions with $\tilde{d}_t$ variables with $\tilde{d}_t:=\sum_{j=1}^t (d_{a,j}+d_{s,j})$. The fact that $\tilde{d}_{t_1}\leq \tilde{d}_{t_2}$ for $t_1\leq t_2$ then implies
that the hypothesis spaces of deep Q-learning vary with $t$.
Under Assumption \ref{Assumption:Markov}, if $d_{a,t}$ and $d_{s,t}$ do not vary with $t$, then   $Q_{t}^*$, $t=1,\dots,T$ are functions with the same number of variables, which leads to the same hypothesis space for all times.
We also present a compactness assumption for the action and state spaces.

\begin{assumption}\label{Assumption:Bounded}
 Assume $\tilde{\mathcal A}_t=[0,1]^{d_{a,t}}$ and $\tilde{\mathcal S}_t=[0,1]^{d_{s,t}}$.
\end{assumption}

Assumption \ref{Assumption:Bounded} can be satisfied by using a standard scaling technique directly, provided that the action spaces and state spaces are compact. This is a mild assumption since data are always collected to be bounded.
Recall that $L_t^2$ is the space of square-integrable functions
with respect to  $P_t$, as defined in (\ref{likelihood}). The following distortion assumption describes the difference between $P_t$ and the Lebesgue measure.
\begin{assumption}\label{Ass:distortion}
For $p\geq 1$, denote  $J_{p,t}$ as the identity mapping:
$
     L^2_t      ~~ {\stackrel{J_p}{\longrightarrow}}~~   L^p([0,1]^{\tilde{d}_t})
$
and $\mathcal J_{p,T}=\max_{t=1,\dots,T}\|J_{p,t}\|,$ where
$\|J\|$ is the spectral norm of the operator $J$.
We assume $\mathcal J_{p,T}<\infty$.
\end{assumption}

\begin{figure}[!t]
\centering
\includegraphics*[scale=0.30]{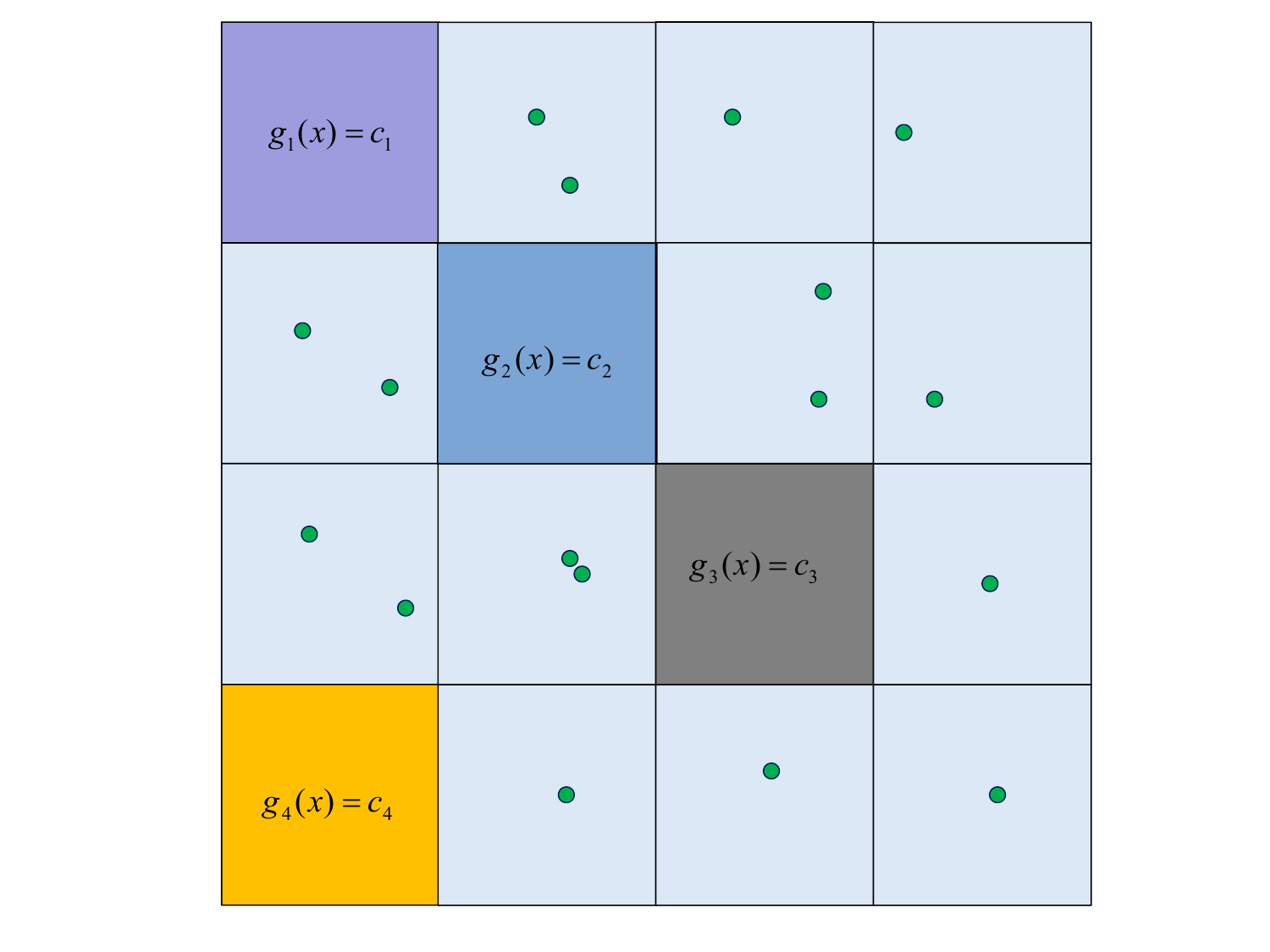}
\hfill \caption{Support of $P_t$ and sparseness of $Q_t^*$}
\label{Fig:support}
\end{figure}

It is obvious that
$\mathcal J_{p,t}$   measures the extent to which $P_t$ distorts the
Lebesgue measure.  Since $Q_t^*$ is frequently spatially sparse, if the support of $P_t$ is out of the support of $Q_t^*$, then all samples are useless in the learning process, as shown in Figure \ref{Fig:support}. Therefore, Assumption \ref{Ass:distortion} is necessary and reasonable.   It holds for all $p\geq2$ when $P_t$ is the uniform distribution.
Then, we present the assumption of spatially sparse and piecewise constant on optimal Q-functions as follows.

\begin{assumption}\label{Ass:piece-constant}
For any  $t=1,\dots,T$,  there exist   $s_t,N_t\in\mathbb N$ such that $Q_t^*\in \mathcal C^{(2U,s_t,N_t^{\tilde{d}_t})}$.
\end{assumption}

As discussed in Sec. 3.1, Assumption \ref{Ass:piece-constant} is standard for numerous applications. As shown in Theorem \ref{Theorem:approx-piece-constant}, each $Q^*_t$ corresponds to a deep net with two hidden layers and $ \mathcal O(N_t^{\tilde{d}_t})$ free parameters.
Denote by $\mathcal H_{N_t,\tau,t}$ the set of deep nets structured as in
 Theorem \ref{Theorem:approx-piece-constant} for $t=1,\dots,T$.   Given the dataset $D=\{\mathcal T_{T,i}\}_{i=1}^m=\{(\mathbf s_{T+1,i},\mathbf a_{T,i},\mathbf R_{T,i})\}_{i=1}^m$, we can deduce Q-functions via $\hat{Q}_{T+1,N_T+1}=0$ and
\begin{equation*}\label{deep Q learning algorithm for Q-constant}
    \begin{split}
    &\hat{Q}_{t,N_t,\tau}(\mathbf s_t,\mathbf a_t)
      = {\arg\min}_{Q_t\in \mathcal H_{N_t,\tau,t}}
       \\&\mathbb E_m \left[\left(R_t+\max_{a_{t+1}}\hat{Q}_{t+1}(\mathbf S_{t+1},\mathbf
     A_{t},a_{t+1})-Q_t(\mathbf S_t,\mathbf
     A_t)\right)^2\right].
    \end{split}
\end{equation*}
Then, the policy  derived from deep Q-learning is defined by
\begin{equation}\label{policy-piececonstant}
  \hat{\pi}_{N,\tau}=\{\hat{\pi}_{1,N_1,\tau},\dots,\hat{\pi}_{T, N_T,\tau}\},
\end{equation}
where
\begin{equation}\label{deep Q learning algorithm for policy-constant}
    \begin{split}
   \hat{\pi}_{t,N_t,\tau}(\mathbf s_t,\mathbf
   a_{t-1})=\arg\max_{a_t\in\tilde{\mathcal A}_t}\hat{Q}_{t,N_t}&(\mathbf s_t,\mathbf
   a_{t-1},a_t), \\&t=1,\dots,T.        
    \end{split}
\end{equation}
Since $N_t$ and $\tilde{d}_t$ vary with   $t$, the network structures at different times are vary. Further noting that  $L=2$ for all $t$, we can parameterize the width in the learning process.  The following assumption on piecewise smooth and spatially sparse is our final assumption.
\begin{assumption}\label{Ass:piece-smooth}
For any  $t=1,\dots,T$,  there exist $r_t>0$, $c_0>0$,  $s_t,N_t\in\mathbb N$ such that $Q_t^*\in Lip^{(r_t,c_0,s_t,N_t^{\tilde{d}_t})}$.
\end{assumption}

The non-differentiability of ReLU results in the necessity of depth in approximating
  functions in $Lip^{(r,c_0)}_{\mathbb I^d}$.
Let $\mathcal H_{t,n_t,N_t,L_t}$ be the set of deep nets structured as in Theorem \ref{Theorem:piecewise-smooth} with respect to $\mathcal Q^*_t$.  We build the hypothesis spaces for Q-learning as  $\tilde{\mathcal Q}_t= H_{t,n_t,N_t,L_t,2U}:=\{f\in H_{t,n_t,N_t,L_t}:\|f\|_{L^\infty}\leq 2U\}$.  Then, Q-functions can be defined by $\hat{Q}_{T,n_{T+1},n_{T+1},N_{T+1},L_{T+1}}=0$ and  for $1\leq t\leq T$,
\begin{equation*}\label{deep Q learning algorithm for Q}
    \begin{split}
      &\hat{Q}_{t,n_t,N_t,L_t}(\mathbf s_t,\mathbf a_t)
      = {\arg\min}_{Q_t\in   \mathcal H_{t,n_t,N_t,L_t,2U}}
       \\&\mathbb E_n\left[\left(R_t+\max_{a_{t+1}}\hat{Q}_{t+1}(\mathbf S_{t+1},\mathbf
     A_{t},a_{t+1})-Q_t(\mathbf S_t,\mathbf
     A_t)\right)^2\right].
    \end{split}
\end{equation*}
The policy derived from Q-learning is defined by
\begin{equation}\label{policy-piecesmooth}
  \hat{\pi}_{n,N,L}=\{\hat{\pi}_{1,n_1,N_1,L_1},\dots,\hat{\pi}_{T, n_T,N_T,L_T}\},
\end{equation}
where
\begin{equation}\label{deep Q learning algorithm for policy}
    \begin{split}
   \hat{\pi}_{t,n_t,N_t,L_t}(\mathbf s_t,\mathbf
   a_{t-1})=\arg\max_{a_t\in\tilde{\mathcal A}_t}&\hat{Q}_{t,n_t,N_t,L_t}(\mathbf s_t,\mathbf
   a_{t-1},a_t), \\&t=1,\dots,T.
   \end{split}
\end{equation}


\subsection{Power of depth in deep Q-learning}
In this subsection, we derive the generalization error for deep Q-learning under the aforementioned assumptions. Our first result shows the power of depth in deep Q-learning when optimal Q-functions are spatially sparse and piecewise constant.

\begin{theorem}\label{Theorem:learning-piece-constant}
  Under Assumptions \ref{Assumption:case-1},  \ref{Assumption:bounded-Reward},  \ref{Assumption:Bounded}, \ref{Ass:distortion},   and  \ref{Ass:piece-constant}, if $\hat{\pi}_{N,\tau}$ is defined by  (\ref{policy-piececonstant})   with
 $\tau=\mathcal O( N^{ \tilde{d}_t-1}s^{-1}m^{-1})$,
then
\begin{equation}
    \begin{split}
    E[&V^*(S_1)-V_{ \hat{\pi}_{N,\tau}}(S_1)]
        \leq
       \\&\hat{C}_1\mathcal J_{p,T}\left(\frac{\log m}{m}\right)^{1/2}
       \\&\cdot \sum_{t=1}^T  \sum_{j=t}^T  (3\mu)^{j-t}    N_t^{\max\{\tilde{d}_{j},\tilde{d}_{j+1}\}/2} (\log N_j)^{1/2},
   \end{split}
\end{equation}
where $\hat{C}_1$ is a constant depending only on $r,p$, and $U$.
\end{theorem}

The use of deep nets to learn discontinuous functions in the framework of supervised learning was first studied in \cite{imaizumi2018deep}. In Theorem \ref{Theorem:learning-piece-constant}, we extend their result from supervised learning to RL  by using the oracle inequality in Theorem \ref{Theorem:oracle}.  Noting that shallow nets \cite{chui1994neural} and linear models \cite{llanas2008constructive} have difficulties in realizing either the spatial sparseness or piecewise-constant property of optimal Q-functions, the corresponding generalization error is worse than the established results in Theorem \ref{Theorem:learning-piece-constant}. As a consequence, traditional Q-learning requires many more samples than deep Q-learning to finish a specific learning task.
This demonstrates the power of depth and explains why deep Q-learning performs so well in numerous applications.
The following corollary shows the generalization error of deep Q-learning when the Markov assumption is imposed.

\begin{corollary}\label{Corollary:learning-piece-constant}
Under Assumptions \ref{Assumption:case-1}-\ref{Ass:piece-constant},    if $d_{a,1}=\dots=d_{a,T}=d_a$, $d_{s,t}=\dots=d_{s,T}=d_s$, $N_1=\dots=N_T=N$, and
  $\hat{\pi}_{N,\tau}$ is defined by  (\ref{policy-piececonstant})   with
 $\tau=\mathcal O( N^{ ({d}_a+d_s)-1}s^{-1}m^{-1})$,
then
\begin{equation}
    \begin{split}
    E[V^*(S_1)-&V_{ {\pi}_{N,\tau}}(S_1)]
        \leq
       \\\hat{C}_1\mathcal J_{p,T}&\left(\frac{N^{d_a+d_s}\log N\log m}{m}\right)^{1/2}
       \\&\cdot\left(\frac{T}{1-3\mu}-\frac{3\mu(1-(3\mu)^T)}{1-3\mu}\right)
   \end{split}
\end{equation}
\end{corollary}

Our next theorem shows the generalization error of deep Q-learning in learning spatially sparse and smooth optimal Q-functions.

\begin{theorem}\label{Theorem:learning-piece-smooth}
  Under Assumptions \ref{Assumption:case-1},  \ref{Assumption:bounded-Reward}, \ref{Assumption:Bounded},  \ref{Ass:distortion} and \ref{Ass:piece-smooth}, if  $\hat{\pi}_{n,N,L}$ is defined by
  (\ref{policy-piecesmooth}) with
$$
      n_t=\left(\frac{ms_t^2}{ N_t^{\max\{\tilde{d}_{t+1},\tilde{d}_t\}+2\tilde{d}_t/p}}\right)^{\frac{\tilde{d}_t}{2r+\tilde{d}_t}}, \quad t=1,\dots,T,
$$
then
\begin{equation*}\label{generalization-error-dql}
    \begin{split}
    E[V^*(S_1)&-V_{\hat{\pi}_{_{n,N,L}}}(S_1)] \leq
    \\& \hat{C}_2\mathcal J_{p,T}\sum_{t=1}^T \sum_{j=t}^T  (3\mu)^{j-t} m^{-\frac{r}{2r+\tilde{d}_j}}s_j^{\frac{\tilde{d}_j}{2r+\tilde{d}_j}}
        \\&\cdot N_j^{\frac{pr\max\{\tilde{d}_{j+1},\tilde{d}_j\}
        -\tilde{d}_j^2}{(2r+\tilde{d_j})p}}(\max\{\tilde{d}_j,\tilde{d}_{j+1}\})^\frac{3}2\log(mN_j),
   \end{split}
\end{equation*}
where $\hat{C}_2$ is a constant depending only on $r,p$, and $U$.
\end{theorem}

Since we consider a general case for deep Q-learning, the generalization error established in Theorem \ref{Theorem:learning-piece-smooth} seems a little bit overly sophisticated, by showing its dependence on the smoothness $r_t$, sparsity $s_t$, number of partitions $N_t$, dimension $\tilde{d}_t$, distortion $\mathcal J_{p,T}$, probability parameter $\mu$, and size of samples $m$. We shall discuss the impact of each factor and simplify our result in the rest of this section.  As shown in Theorem \ref{Theorem:piecewise-smooth}, obtaining an approximation accuracy of order $\mathcal O(n^{-r/d}sN^{-d/p})$ requires a number of free parameters equal to at least $\mathcal O(nN^d)$. This shows a bias-variance trade-off by noting that the capacity of deep nets depends heavily on the number of free parameters.
Under these conditions,  $n_t$ in Theorem \ref{Theorem:learning-piece-constant}
is selected to balance the bias and variance. Since the reward functions in Q-learning are practically extremely sparse,
the sparsity $s_t$, compared with the number of partitions $N_t^d$, is often extremely small, which together with $pr\leq \min_{1\leq j\leq T}\tilde{d}_j$   yields very good generalization error estimates for deep Q-learning. In the following, we present a corollary for Theorem \ref{Theorem:learning-piece-smooth} under added assumptions to explicitly exhibit the generalization error.

\begin{corollary}\label{Corollary:learning-piece-smooth}
Under Assumptions \ref{Assumption:case-1}-\ref{Ass:distortion}  with $p=2$ and Assumption \ref{Ass:piece-smooth},  if $d_{a,1}=\dots=d_{a,T}=d_a$, $d_{s,t}=\dots=d_{s,T}=d_s$, $s_1=\dots=s_T=s$, $r_1=\dots=r_T$, $N_1=\dots=N_T=N$,
 and
 $$
      n=\left(\frac{ms^2}{ N^{2d_a+2d_s}}\right)^{\frac{d_a+d_s}{2r+d_a+d_s}},
$$
then
\begin{equation*}
    \begin{split}
    E[V^*(&S_1)-V_{ \hat{\pi}_{_{n,N,L}}}(S_1)]\leq
    \\&\hat{C}_3m^{-\frac{r}{2r+d_a+d_s}}\log (mN) s^{\frac{d_a+d_s}{2r+d_a+d_s}}     
    \\&\cdot N^{\frac{(2r-d_a-d_s)(d_a+d_s)}{4r+2d_a+2d_s}}\sum_{t=1}^T   \left(\frac{T}{1-3\mu}-\frac{3\mu(1-(3\mu)^T)}{1-3\mu}\right),
   \end{split}
\end{equation*}
where $\tilde{C}_3$ is a constant independent of $N,s,m$, or $T$.
\end{corollary}

From Corollary \ref{Corollary:learning-piece-smooth}, a generalization error bound of order
$\mathcal O(m^{-\frac{r}{2r+d_a+d_s}}N^{\frac{(2r-d_a-d_s)(d_a+d_s)}{4r+2d_a+2d_s}}s^{\frac{d_a+d_s}{2r+d_a+d_s}})$ is derived. If $r$ is large, then the dominant term is $m^{-\frac{r}{2r+d_a+d_s}}$. Under this circumstance, numerous data are required to produce a high-quality policy, just as AlphaGo did in \cite{silver2016mastering}. If $r$ is relatively small, then $N^{\frac{(2r-d_a-d_s)(d_a+d_s)}{4r+2d_a+2d_s}}s^{\frac{d_a+d_s}{2r+d_a+d_s}} $ is the dominant term, implying that there are a few candidates available to make a decision, which is also popular in practice \cite{zheng2018drn}. Furthermore, given that the linear models and shallow nets cannot approximate the spatially sparse and piecewise smooth functions well \cite{chui1994neural,petersen2018optimal}, it is difficult to derive a similar result for traditional Q-learning as Theorem \ref{Theorem:learning-piece-constant} and Corollary \ref{Corollary:learning-piece-smooth}, which
  demonstrates the power of depth in deep Q-learning.

To end this section, we differentiate our results are from those of \cite{fan2020theoretical}, where theoretical verification is also conducted for deep Q-learning. As discussed in Sec. 1.3, the setting in \cite{fan2020theoretical} is available for RL with infinite horizons and requires strong assumptions on the optimal Q-functions and likelihood  $P$ (defined by (\ref{likelihood under P})),  which are difficult to check in practice. As shown in Sec. 3.1, Assumptions 2, 5, 6, and 7 on $Q_t^*$ are easily satisfied for numerous real-world applications (e.g., Table \ref{Table:Example}). Furthermore, we only impose two extra assumptions including Assumptions \ref{Assumption:case-1} and \ref{Ass:distortion} on the likelihood $P$, which is essentially looser than the concentration coefficient assumption in \cite{fan2020theoretical} and easily satisfied in practice. The weakness of the assumptions and availability in practice are main the reasons that our derived generalization error bounds behave exponentially badly in the time horizon. It would be interesting to find more suitable assumptions from real-world applications to reduce this negative effect.

\section{Experimental Results}\label{Sec.Experiment}

In this section, we apply deep Q-learning to the beer game, a well-known supply chain management problem, and a recommender system application, to illustrate the roles of the depth of neural networks, rewards, and data size in RL.

\subsection{Beer Game Experiment}
The first experiment is conducted in the context of an inventory management problem, named the Beer Game. Beer Game is a multi-agent supply chain management problem, with four agents participating, from upstream to downstream of which are the manufacturer, distributor, warehouse and retailer. In each period of the game, each agent observes the current inventory state and decides on an order to send to his predecessor (supplier). We examine how DQN can help agents decide on the right orders to minimize the total inventory in hand over a long time. The detailed introduction as well as an RL framework of the beer game can be found in Section 4 of the supplementary material. We report our experiment designs and numerical results in the following subsections. Each subsection contains experiments based on simulations and three real-world data sets. The experimental settings based on the simulated data and real-world data sets are given in Section 5 and Section 6 of the supplementary material. 

\subsubsection{Power of the depth}

Our first simulation focuses on the power of depth in deep Q-learning. According to Theorems \ref{Theorem:learning-piece-constant} and  \ref{Theorem:learning-piece-smooth}, the effect of depth depends heavily on the reward property. Therefore, we use the shaped reward in \cite{oroojlooyjadid2022deep} in this simulation and record our method as shaped reward deep Q-networks (SRDQN). The based stock policy (bs for short) is regarded as a baseline. As there are four agents in the beer game, we only apply deep Q-learning on the first agent while applying bs on the three remaining agents. In this way, we record our approach as shaped reward deep Q-networks and based stock policy (SRDQN-BS). For further clarity, we refer the readers to Section 4 and Section 5 of the supplementary material for details.

\begin{figure}[!t]
	\centering
	\includegraphics*[scale=0.30]{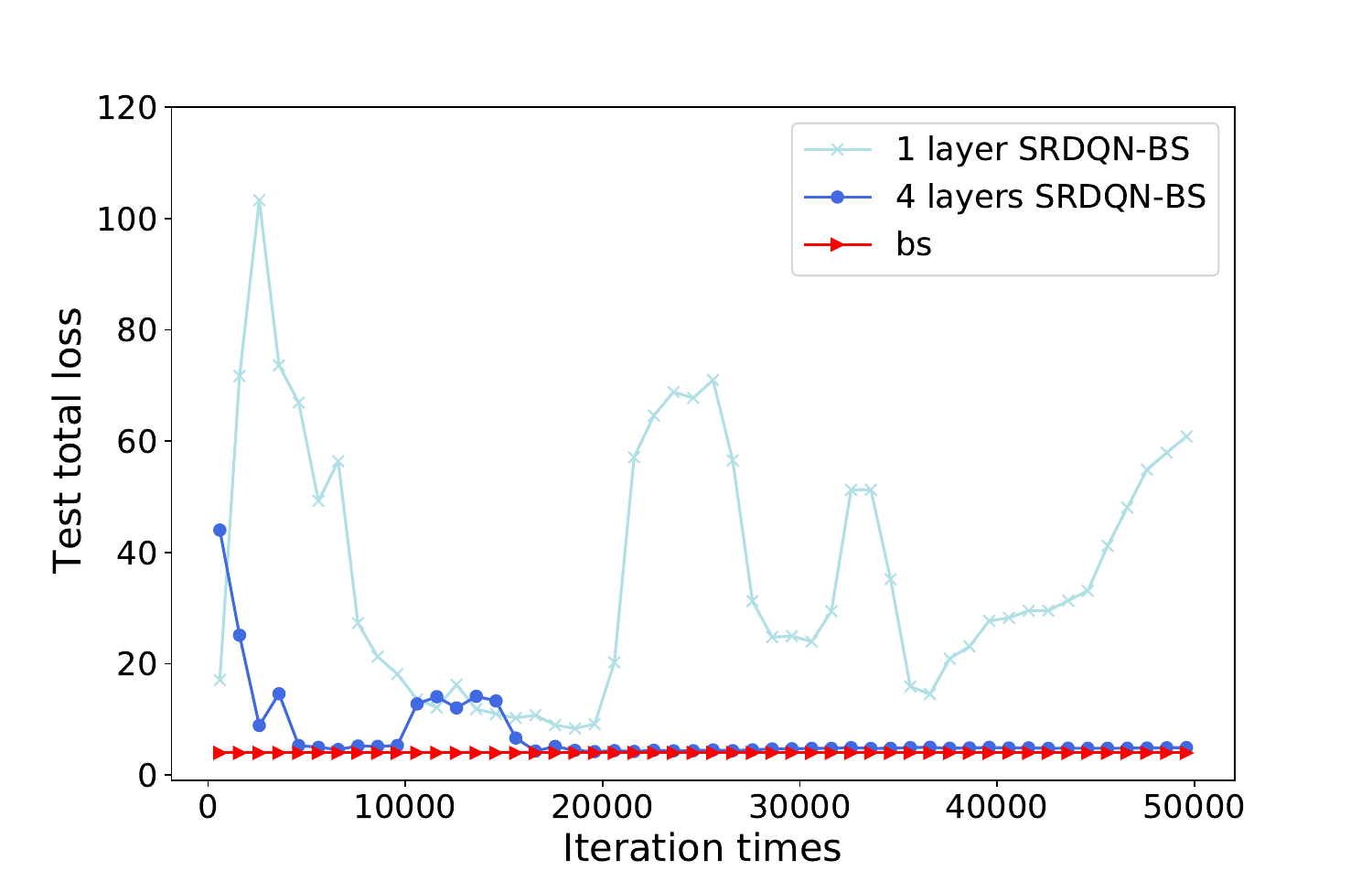}
	\hfill \caption{Performance of SRDQN-BS policy with different-depth-SRDQN }
	\label{53}
\end{figure}

As shown in Section 4 of the supplementary material, the reward, as a function of actions, possesses the spatially sparse and piecewise constancy property. Our theoretical assertions suggest that deep learning outperforms the classical (shallow) approach in such an RL problem. To verify this, we test the SRDQN-BS policy with five cases, with one, two, three, four, and five hidden layers in the SRDQN. Results of the one- and four-layer cases are shown in Figure \ref{53}.

From Figure \ref{53}, there are three interesting observations: (1) The test total loss of the four-layers SRDQN-BS is much less than that of the one-layer SRDQN-BS, showing that
deepening the network is crucial for improving the performance of the classical shallow learning policy. (2) After a few iterations, the prediction of four-layer SRDQN-BS stabilizes, showing that it can generalize well, which is beyond the capability of the one-layer SRDQN-BS. This verifies Theorem \ref{Theorem:learning-piece-constant} in the sense that the variance of deep Q-learning is not large,  since deepening the network does not essentially enlarge the capacity of hypothesis space. (3) After 15000 iterations, four-layer SRDQN-BS performs almost as well as bs, showing that our adopted approach can achieve an almost optimal generalization performance. All these observations show that depth plays an important role in deep Q-learning if spatially sparse and piecewise constant rewards are utilized.

\begin{figure}[!t]
	\centering
	\includegraphics*[scale=0.30]{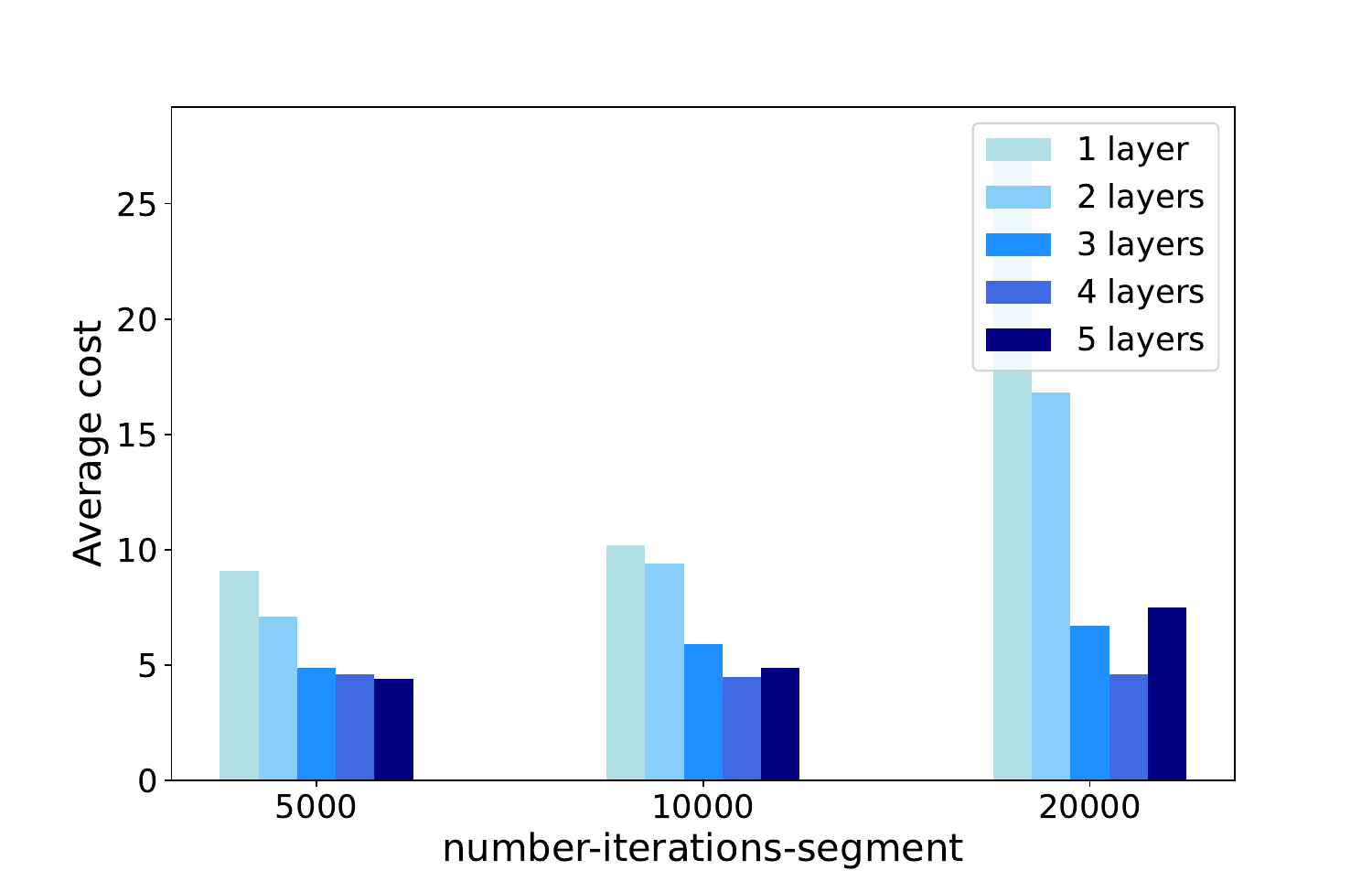}
	\hfill \caption{Best-performance-segment of SRDQN-BS policy with different iteration ranges}
	\label{53segment}
\end{figure}

To show the power of depth and stability of SRDQN-BS with different layers, we also extract the best-performance-segment of SRDQN-BS with consecutive iterations 5000, 10000 and 20000 of five cases,  as compared in Figure \ref{53segment}.

\begin{figure*}[!]
	\centering
	\includegraphics*[scale=0.35]{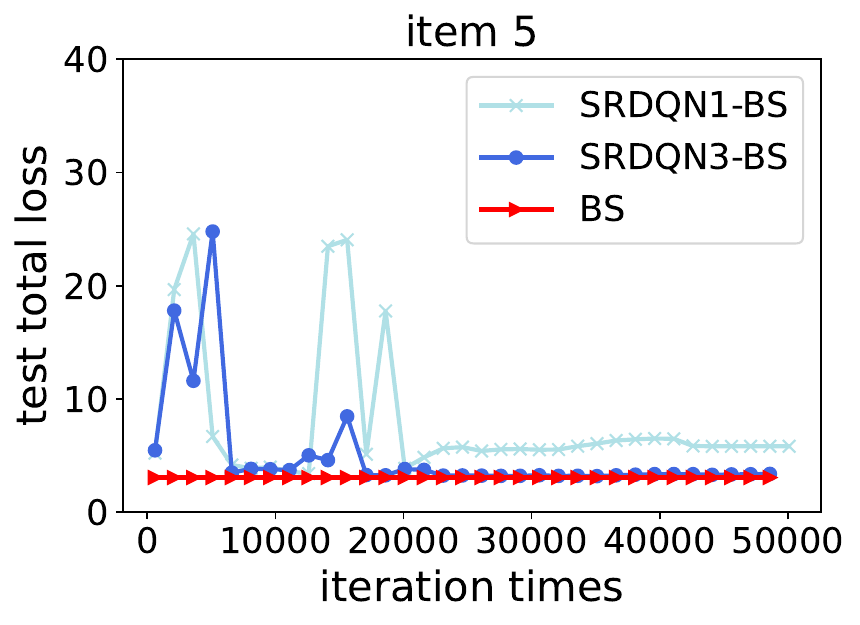}
	\includegraphics*[scale=0.35]{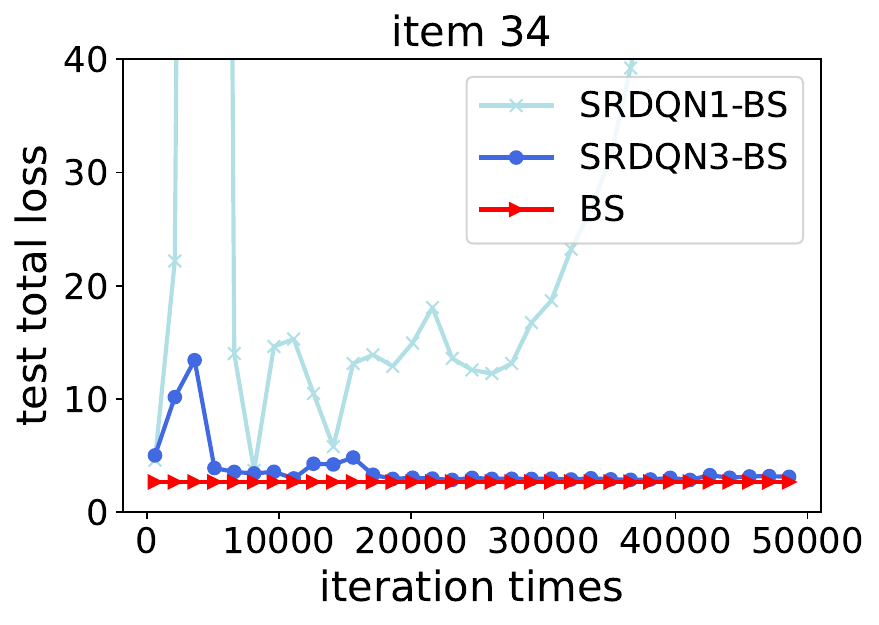}
	\includegraphics*[scale=0.35]{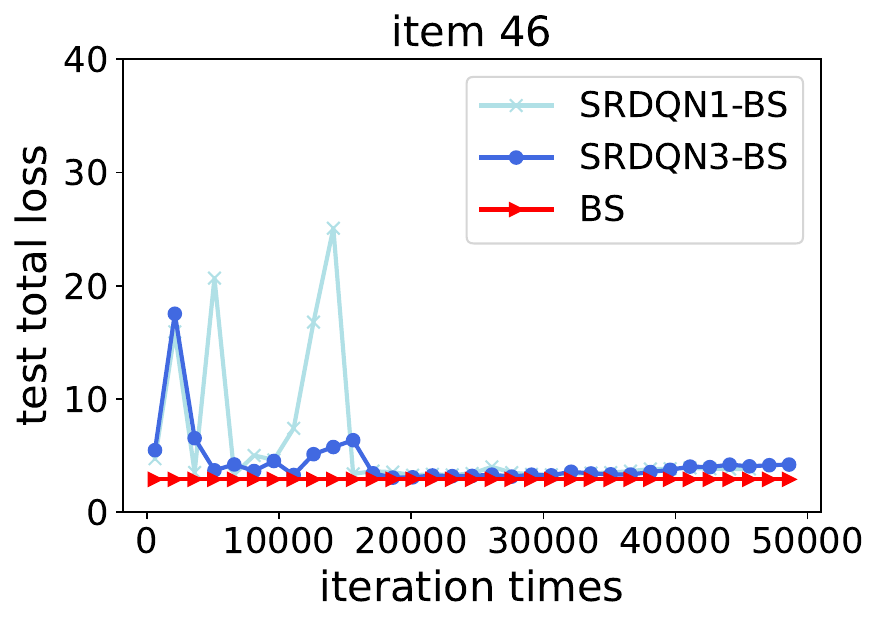}
	\hfill \caption{Performance of SRDQN-BS policy with different-depth-SRDQN}
	\label{real data SRDQN-BS}
\end{figure*}

\begin{figure*}[!]
	\centering
	\includegraphics*[scale=0.35]{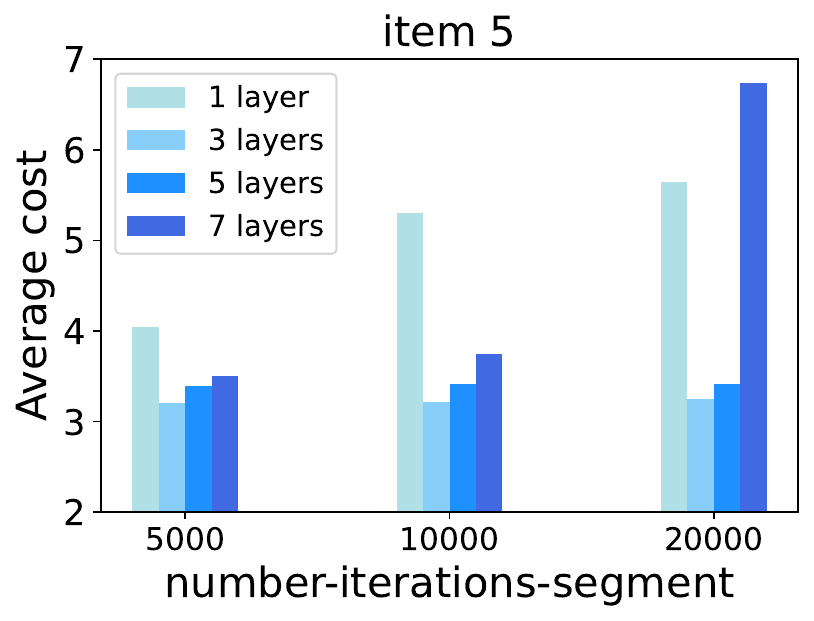}
	\includegraphics*[scale=0.35]{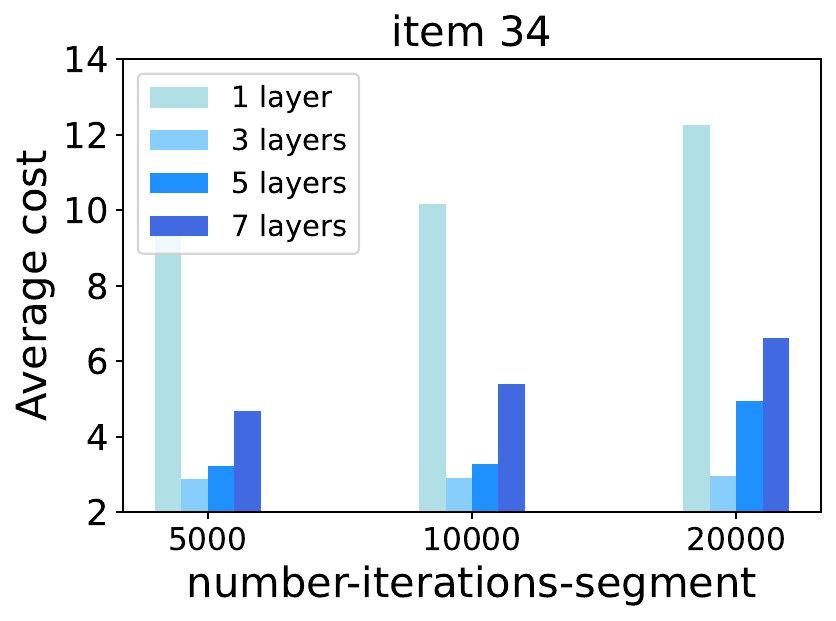}
	\includegraphics*[scale=0.35]{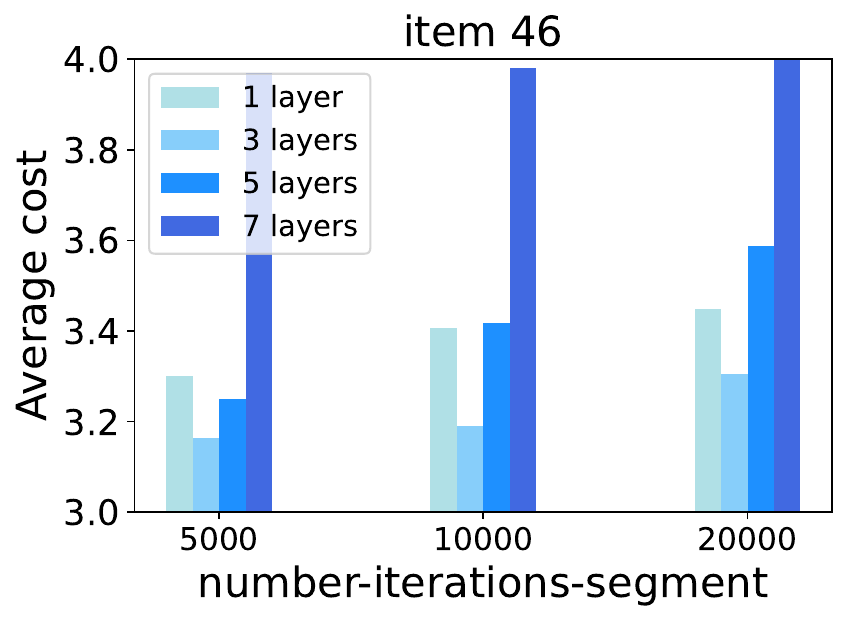}
	\hfill \caption{Best-performance-segment of SRDQN-BS policy with different iteration ranges}
	\label{real data Best-performance-segment}
\end{figure*}

Two interesting phenomena are exhibited in Figure \ref{53segment}: (1) Before a threshold value (L=4 in this simulation), depth plays a positive role in SRDQN-BS. This shows again the power of depth in deep Q-learning. (2) After the threshold value, depth is not so important in generalization,  since five-layer SRDQN-BS performs a little bit worse than the four-layer policy. This does not contradict our theoretical assertions. In fact, according to our proofs in the supplementary material, the constants $\hat{C}_1$ and $\hat{C}_2$ depend on the number of layers, which may cause a small oscillation. Furthermore, the covering number estimate (Lemma \ref{Lemma:covering-number}) shows that the capacity of deep nets increases as they are deepened, leading to an instability phenomenon for five-layer  SRDQN-BS.

We further conduct experiments based on real-world historical demand data for three different items to examine the power of depth in deep Q-learning. The training process of one- and three-layer SRDQN-BS policy is shown in Figure \ref{real data SRDQN-BS}, where the bs policy acts as the optimal baseline.

The findings are quite similar to those in synthetic simulations. It's clear that depth is essential for good performance of SRDQN. In all three experiments, SRDQN with 3 layers converges to a stable point with a small total loss more quickly than SRDQN with 1 layer. In the experiment of item 34, SRDQN with 1 layer even doesn't converge in the end. The second point that deserves to be noticed is that the convergence point of SRDQN-BS policy with 3 layers is pretty close to bs policy in all three experiments, which indicates that it achieves near-optimal performance. This shows the strong generalization ability of the SRDQN-BS policy with 3 layers.

We conduct a more comprehensive comparison by extracting the performance of SRDQN-BS at different training iterations. We implement SRDQN-BS with 1 layer, 3 layers, 5 layers, and 7 layers. The results are shown in Figure \ref{real data Best-performance-segment}.

Clearly, the SRDQN-BS policy with 1 layer performs badly while the SRDQN-BS policy with 3 layers performs quite well. However, continuing to increase the depths in SRDQN, such as increasing to 5 and 7 layers, doesn't keep improving performance but worsens the performance. This is because deeper nets lead to instability in the training process.

\subsubsection{Influence of different rewards}
\begin{figure*}[!t]
\begin{minipage}[b]{0.49\linewidth}
\centering
\includegraphics*[width=8cm,height=5cm]{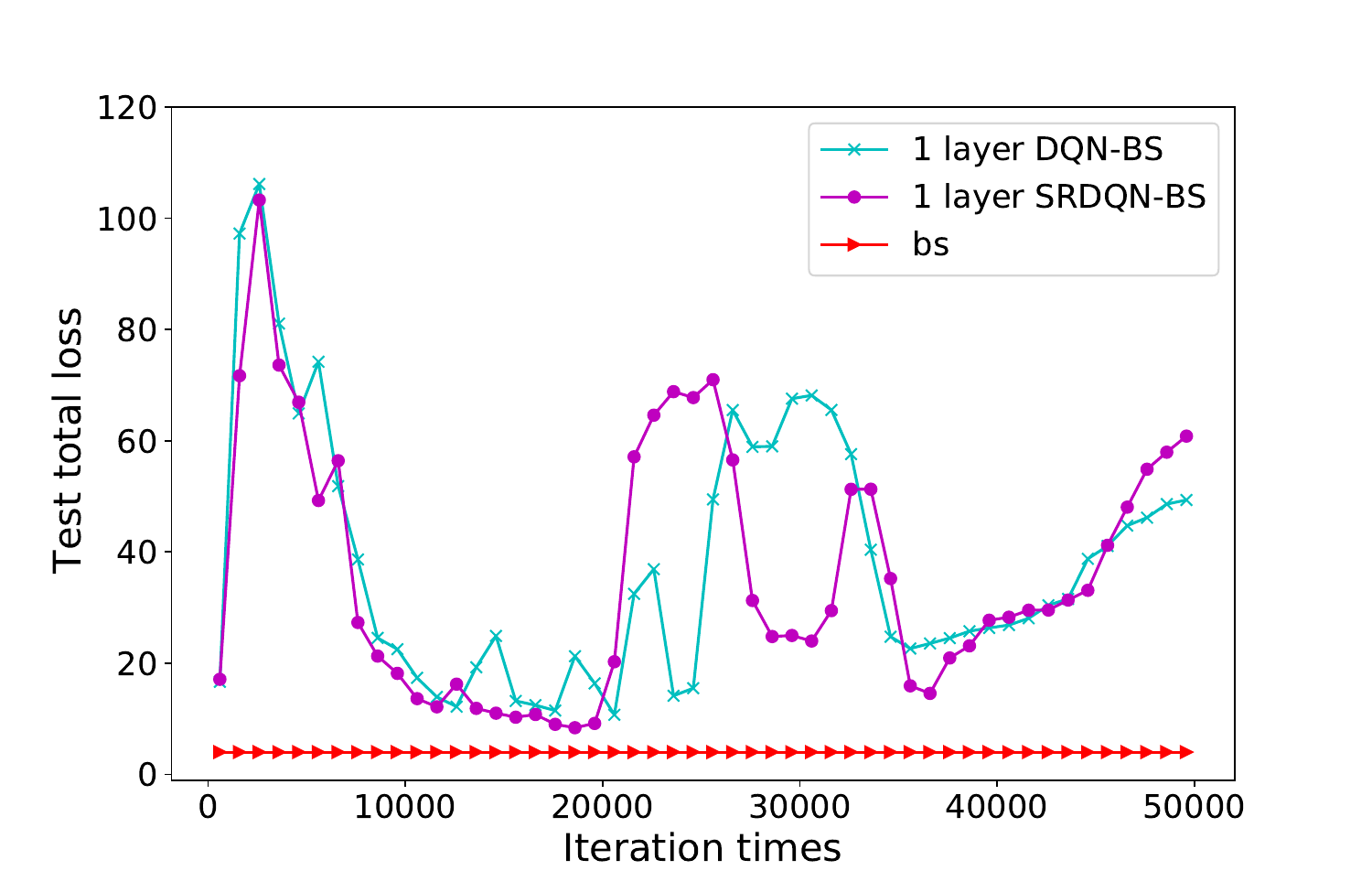}
\centerline{{\small (a) shallow Q-learning}}
\end{minipage}
\hfill
\begin{minipage}[b]{0.49\linewidth}
\centering
\includegraphics*[width=8cm,height=5cm]{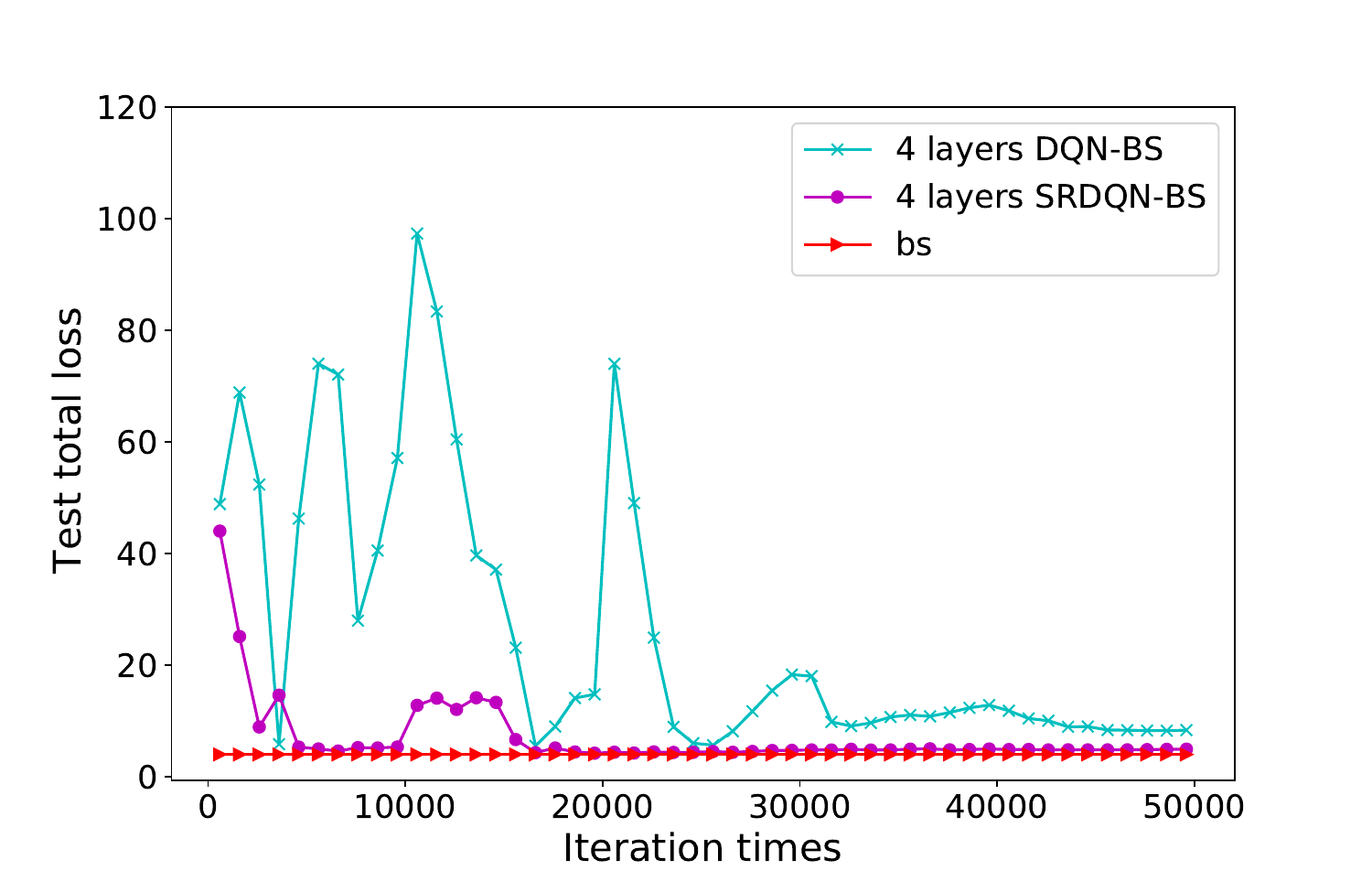}
\centerline{{\small   (d)   deep Q-learning}}
\end{minipage}
\hfill
\caption{Comparison of DQN-BS policy and SRDQN-BS policy with different depth}
\label{54}
\end{figure*}

In this simulation, we shall show the important role of reward in deep Q-learning. We use the classical rewards in the beer game (see Section 4 of the supplementary material) rather than the shaped one proposed in \cite{oroojlooyjadid2022deep} and then obtain the deep Q-learning–bs (DQN–BS) policy. DQN-BS and SRDQN-BS differ only in their rewards.    Comparison of DQN-BS and SRDQN-BS policy with one-layer and four-layers are shown in Figure \ref{54}.

\begin{figure*}[!]
	\centering
	\includegraphics*[scale=0.35]{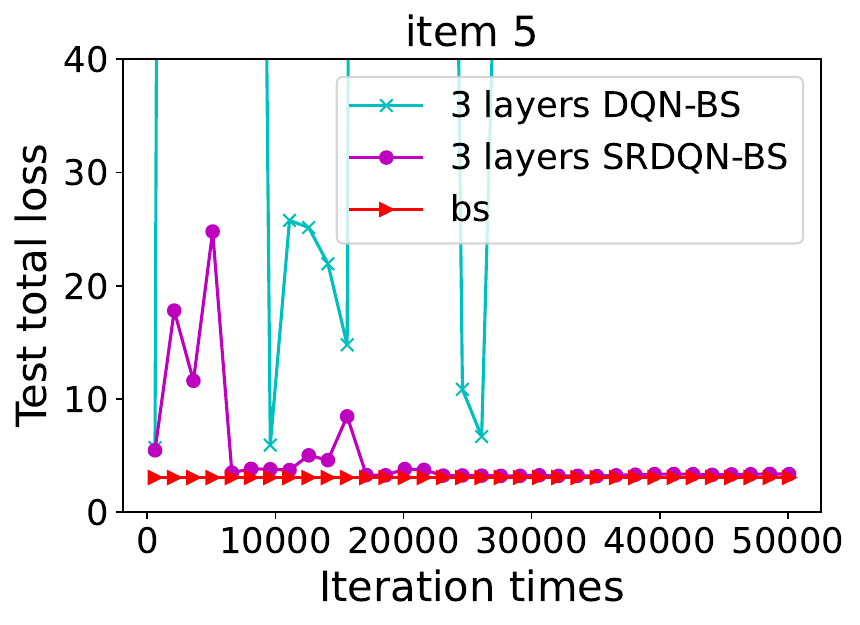}
	\includegraphics*[scale=0.35]{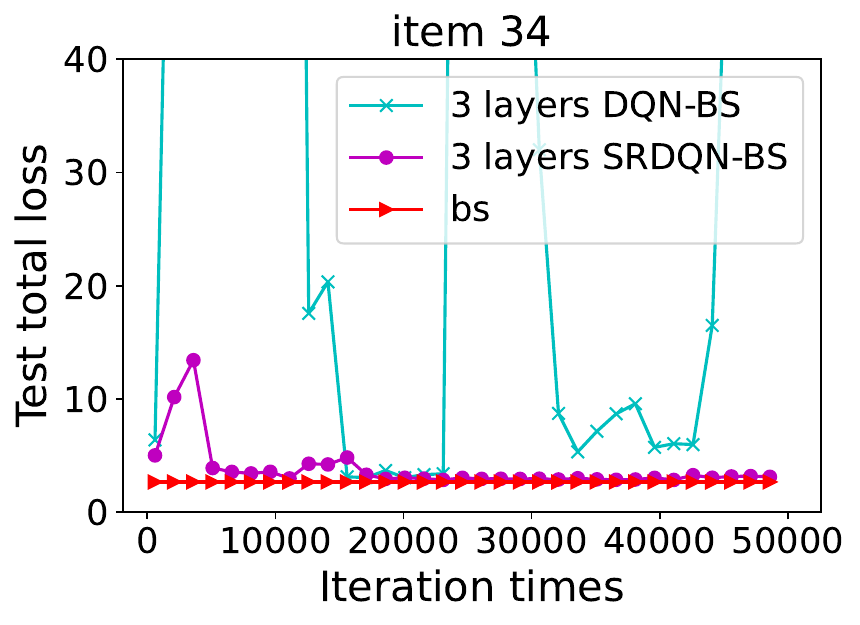}
	\includegraphics*[scale=0.35]{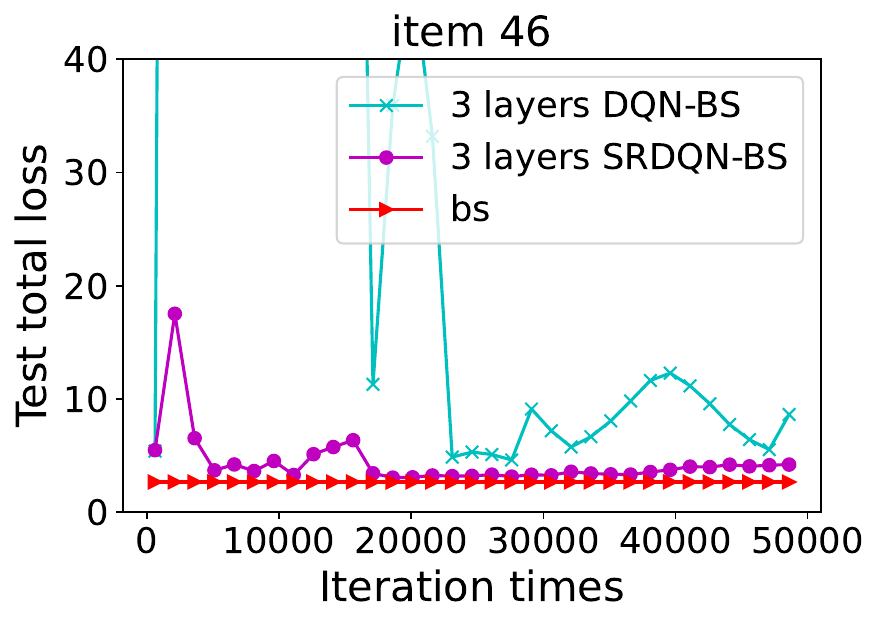}
	\hfill \caption{Comparison of DQN-BS policy and SRDQN-BS policy with different depth}
	\label{real data Comparison}
\end{figure*}

\begin{figure*}[!]
	\centering
	\includegraphics*[scale=0.35]{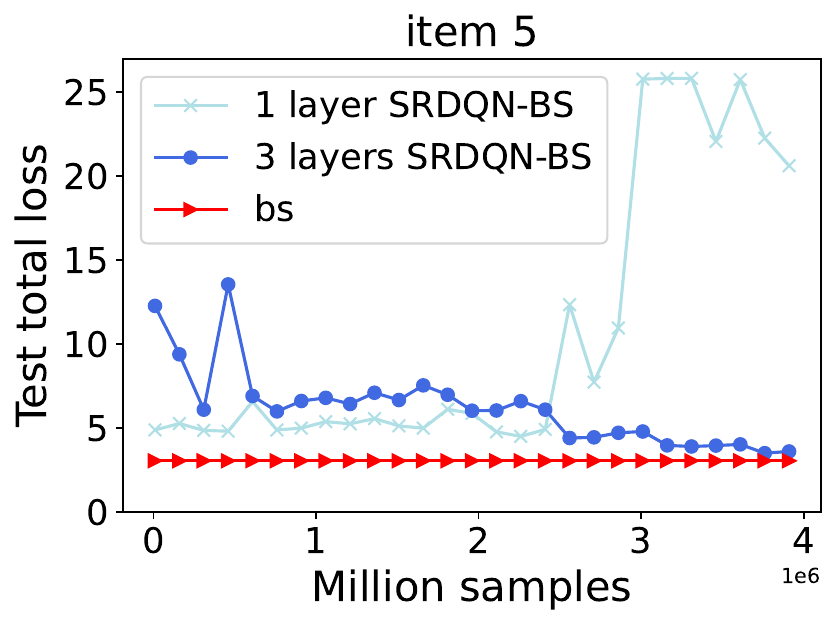}
	\includegraphics*[scale=0.35]{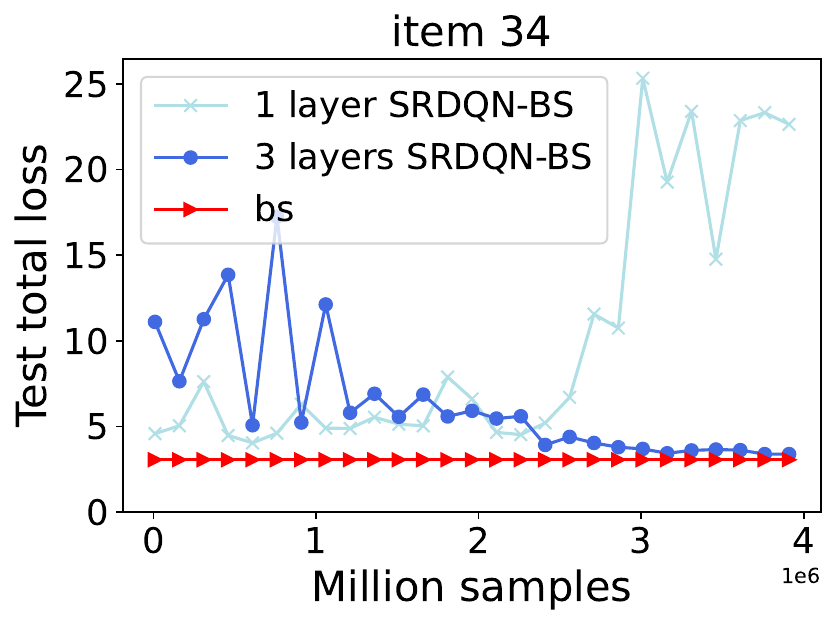}
	\includegraphics*[scale=0.35]{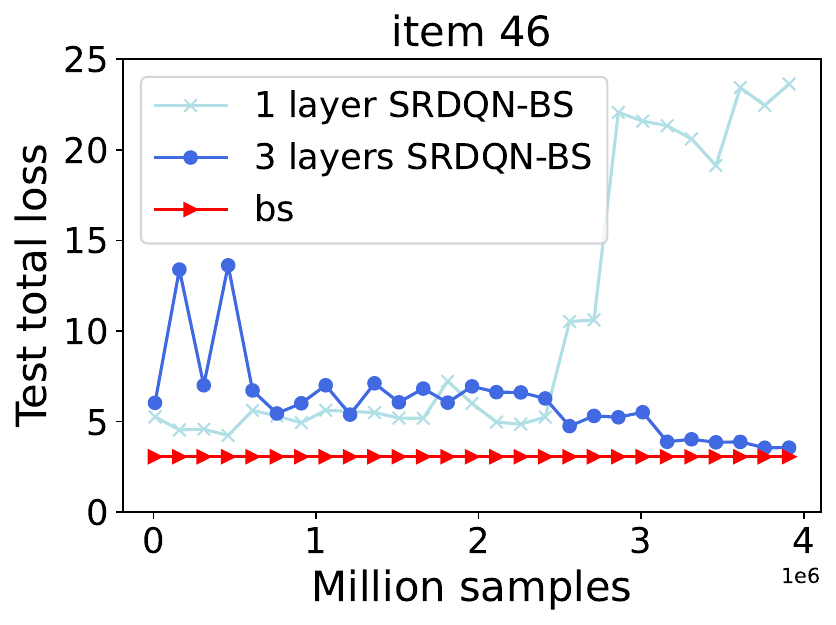}
	\hfill \caption{Role of data size in SRDQN-BS}
	\label{real data size}
\end{figure*}

 Figure \ref{54} exhibits three important findings: (1) From Figure \ref{54} (a), if the hypothesis space is not appropriately
selected, then rewards do not affect the performance of Q-learning substantially. This shows the necessity of taking different hypothesis spaces in Q-learning and implies the importance of Theorem \ref{Theorem:oracle}. (2) From Figure \ref{54} (b), when the suitable hypothesis space is used, then rewards are key to improving the performance of Q-learning. The perfect performance of SRDQN-BS is based on suitable rewards and hypothesis space, which verifies Theorems \ref{Theorem:approx-piece-constant} and \ref{Theorem:piecewise-smooth}; (3) As shown in Section 4 of the supplementary material, the rewards of DQN-BS are also spatially sparse and piecewise constant. As compared with the rewards of SRDQN-BS, the only difference is that there are many more pieces in DQN-BS. According to Theorems \ref{Theorem:learning-piece-constant} and \ref{Theorem:learning-piece-smooth}, the generalization error of DQN-BS should be worse than that of SRDQN-BS. This assertion is verified by comparing Figures \ref{54} (a) and \ref{54} (b). In particular, deep Q-learning improves the performance of shallow Q-learning significantly for SRDQN-BS, while conservatively for DQN-BS.

We report the comparison of SRDQN-BS policy and DQN-BS policy on three real-world datasets in Figure \ref{real data Comparison}. 
We can see that even with 3 layers, the DQN-BS policy can hardly converge, and it performs much worse than the SRDQN-BS policy with 3 layers. This indicates that reward with nice properties is crucial for the power of deep Q-learning.

\subsubsection{Influence of data size }
\begin{figure}[!t]
	\centering
	\includegraphics*[scale=0.30]{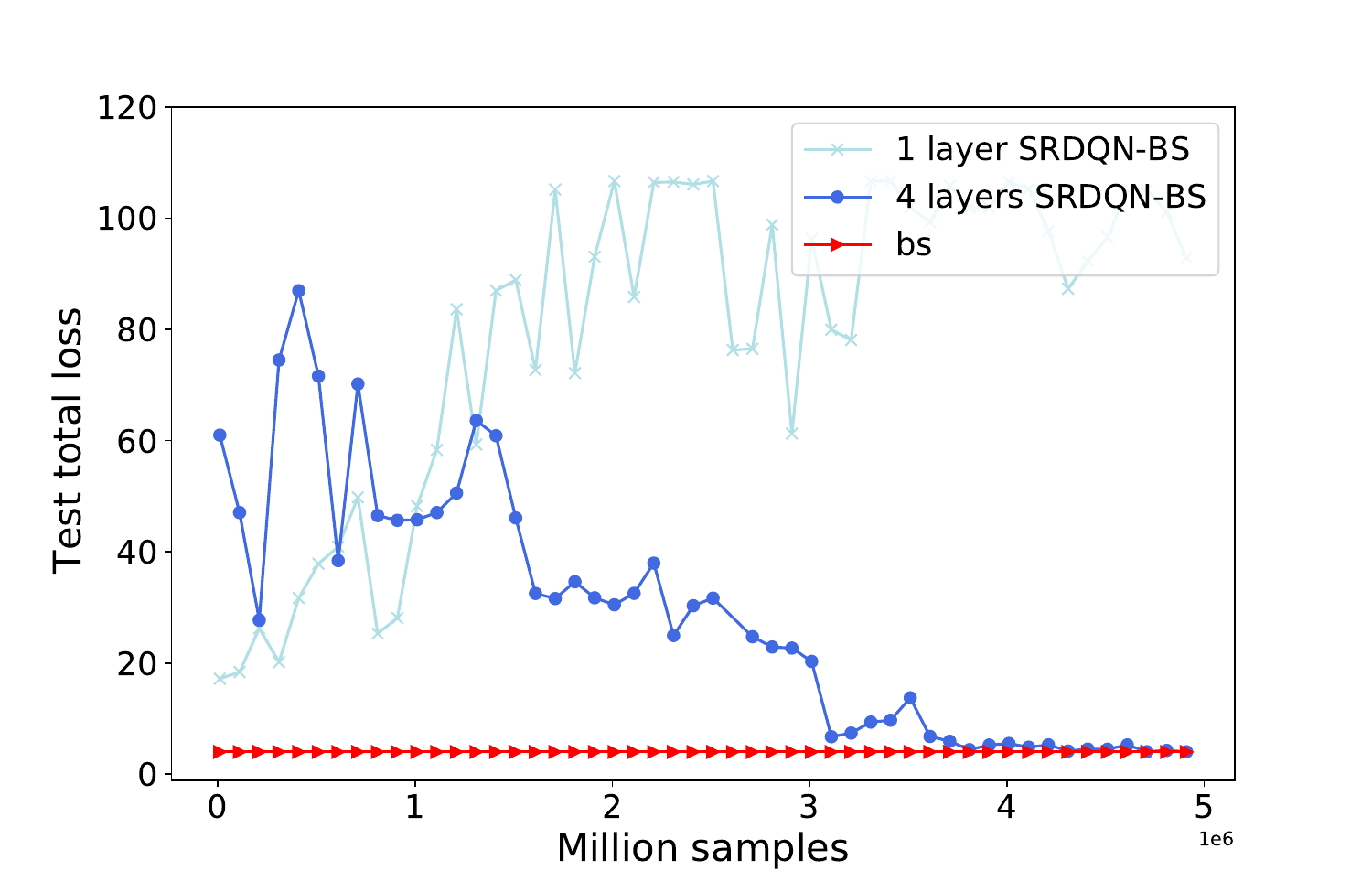}
	\hfill \caption{Role of data size in   SRDQN-BS}
	\label{55}
\end{figure}

In our last simulation, we devoted ourselves to studying the role of data size in deep Q-learning by
investigating how the performance of SRDQN changes with respect to data size. Unlike in the above
experiments that draw the sample with replacement, we draw the sample without replacement to show the role of data size. Specifically, we sample 16 examples from replay memory and discard them after training. By doing this, we relate the number of iterations with the number of samples. After 500 iterations, meeting the minimum experience replay size to start training, 100 samples are used to train SRDQN per iteration. We test the SRDQN-BS policy with two cases (one and four hidden layers of neural networks). The simulation results are shown in Figure \ref{55}.

Based on Figure \ref{55}, we can draw the following three conclusions: (1) As shown in Theorem \ref{Theorem:oracle}, the required size of the sample to guarantee the generalization performance of Q-learning behaves exponentially with $T$, which implies that SRDQN-BS needs large samples.  This demonstrates the poor performance of SRDQN-BS when the data size is smaller than 1,500,000. (2) According to Theorem \ref{Theorem:learning-piece-constant}, compared with shallow Q-learning,  deep Q-learning requires fewer samples to achieve the same long-term rewards. Therefore, the increasing number of samples gradually offsets the negative effects of long-term Q-learning. As a result, the test total cost begins to decrease with respect to the number of samples, as the data size interval changes from 1,500,000 to 3,500,000. However, shallow Q-learning still oscillates because it requires much more samples. (3) After the training of 3,500,000 samples, the four-layer SRDQN-BS  converges to a stable state and performs almost the same as the bs, showing that this policy can achieve the optimal generalization performance, and 3,500,000 is almost the smallest data size needed to guarantee the optimal generalization performance of deep Q-learning in this beer game.

We test the SRDQN-BS policy with 1 layer and 3 layers on three real-world datasets. The results are shown in Figure \ref{real data size}.
The result shows that the SRDQN-BS policy with 3 layers converges to near-optimal after approximately 3 million samples. However, shallow nets diverge after certain periods of training, and it can never achieve near-optimal performance.

\subsection{Recommender System Experiment}
We conduct a second experiment in the context of the recommender system. We look forward to providing criteria for choosing and designing Q-learning frameworks in recommender systems by answering the three questions that we are interested in. Different RL algorithms have been applied to designing recommender systems \cite{ie2019reinforcement,afsar2022reinforcement}, due to the long-term reward maximization nature of reinforcement learning. 
We conduct experiments on a simulation recommender system platform called Recsim \cite{ie2019recsim}. In this simulated recommender system, we use DQN to recommend a set of items for a certain user at each period over a long time. We aim to maximize long-term user engagement, considering user interest shifts. We examine DQN with different depths, with expected reward, which possesses the spatially sparse and piecewise constancy property, and standard reward. The detailed recommender system context that we consider and the RL framework is introduced in Section 7 of the supplementary material. In the following, we describe how to conduct experiments to investigate the aforementioned three different points and how the experimental results verify our theoretical results. 

\begin{figure}[!t]
	\centering
	\includegraphics*[scale=0.5]{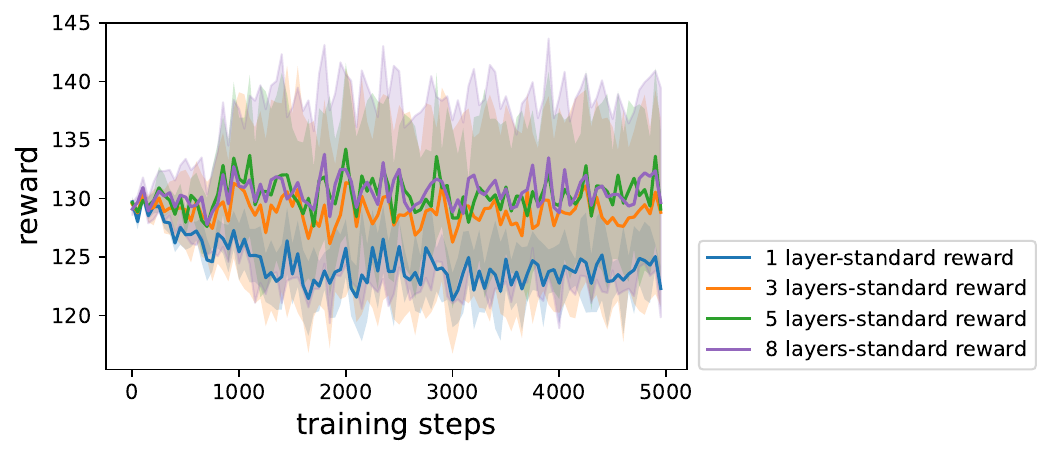}
	\hfill \caption{Performance of DQN with different depths and standard reward}
	\label{standard reward}
\end{figure}

Firstly, we check the power of depth in DQN with the standard reward (DQN-s). The result is shown in Figure \ref{standard reward}. Clearly, DQN-s with 1 layer perform worst. Although DQN-s with 3 layers, 5 layers, and 8 layers are better than DQN-s with 1 layer, they can't achieve an obvious improvement in the training process. This indicates that DQNs can't learn a proper recommendation policy even with a deep net.

\begin{figure}[!t]
	\centering
	\includegraphics*[scale=0.5]{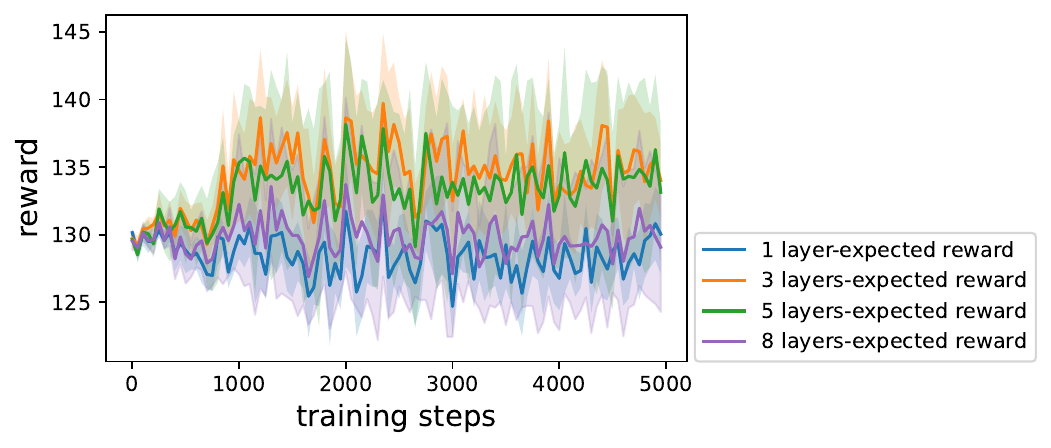}
	\hfill \caption{Performance of DQN with different depths and expected reward}
	\label{expected reward}
\end{figure}

Next, we replace the standard reward with the expected reward, leading to the DQN with the expected reward (DQN-e). The result is shown in Figure \ref{expected reward}. We can see that DQN-e with 1 layer still performs badly. But when the net becomes deeper, i.e., DQN-e with 3 layers and 5 layers, the performance improves obviously after around 1000 training steps. This shows that DQN-e with a proper deep net can perform well in this recommendation task, and so the power of depth in DQN has been shown. On the other hand, when it comes to DQN-e with 8 layers, the performance decreases. This reveals the trade-off between capacity and stability of the deep nets.   

\begin{figure}[!t]
	\centering
	\includegraphics*[scale=0.5]{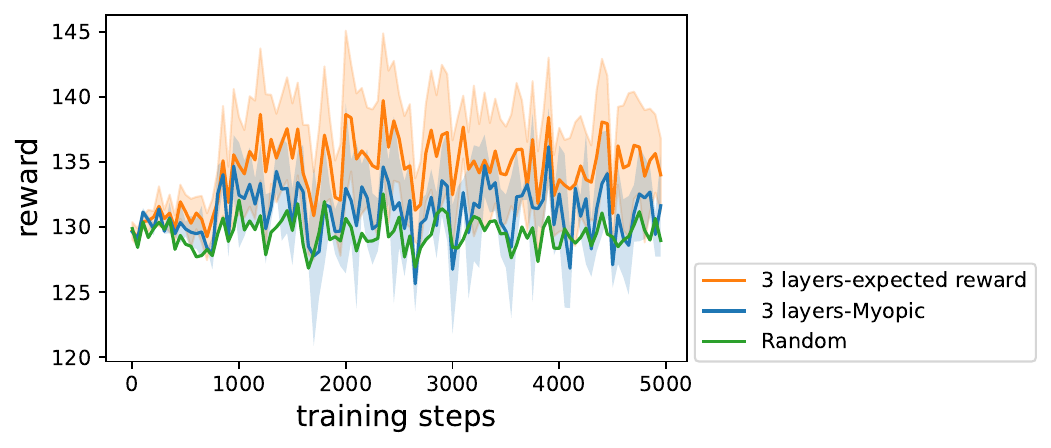}
	\hfill \caption{Comparison of benchmarks and DQN with expected reward}
	\label{benchmark}
\end{figure}

In order to show the effectiveness of the learned policy, we compare DQN-e with 3 layers with two benchmarks. The first benchmark is the Random policy, which randomly samples two documents for recommendation each time $t$. The second benchmark is called Myopic, which means that we train a net without considering long-term reward maximization. Specifically, when we train the DQN net, we update the net in the following form:
\begin{equation*}
    \begin{split}
    Q^{(t)}(s, A) &\leftarrow \alpha^{(t)}\left[r+\max _{A^{\prime}} \gamma Q^{(t-1)}\left(s^{\prime}, A^{\prime}\right)\right]
    \\&+\left(1-\alpha^{(t)}\right) Q^{(t-1)}(s, A)
    \end{split}
\end{equation*}
In Myopic, we set $\gamma$ as 0 to optimize only the immediate reward. We set the net in Myopic policy as the same net in DQN-e. The result is shown in Figure \ref{benchmark}. We can see that the mean reward of the Random policy is approximately a horizontal line. The Myopic policy is better than the Random policy but worse than DQN-e, which shows the importance of considering state transitions and long-term reward maximization. 

\begin{figure}[!t]
	\centering
	\includegraphics*[scale=0.5]{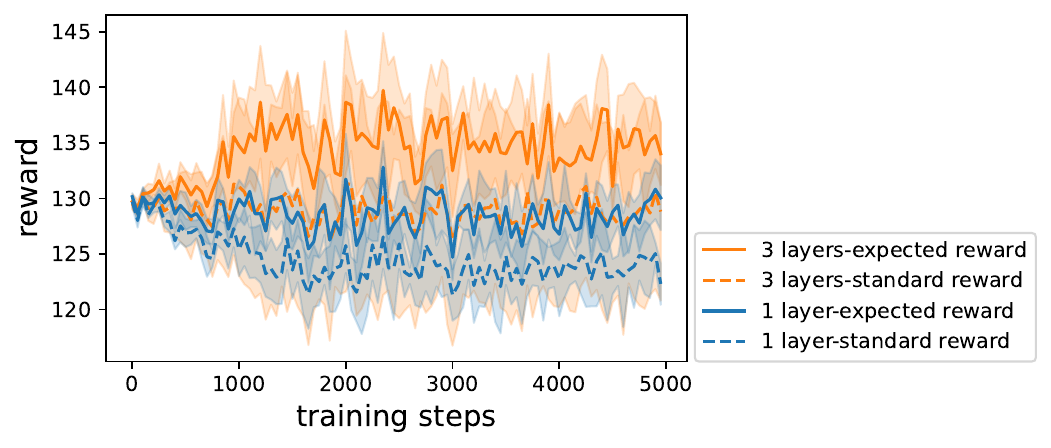}
	\hfill \caption{comparison between DQN with standard reward and DQN with expected reward}
	\label{comparison}
\end{figure}

We compare the performance of DQN-s and DQN-e with different layers in Figure \ref{comparison}. We can reveal the power of DQN only with both proper reward function and proper depth. The management implications here are that DQN is not almighty without any precondition. The first thing we must decide is a reward function with the nice property we proposed in our theory. The second is to decide on a proper depth to uncover the full ability of the DQN method.

\begin{figure}[!t]
	\centering
	\includegraphics*[scale=0.5]{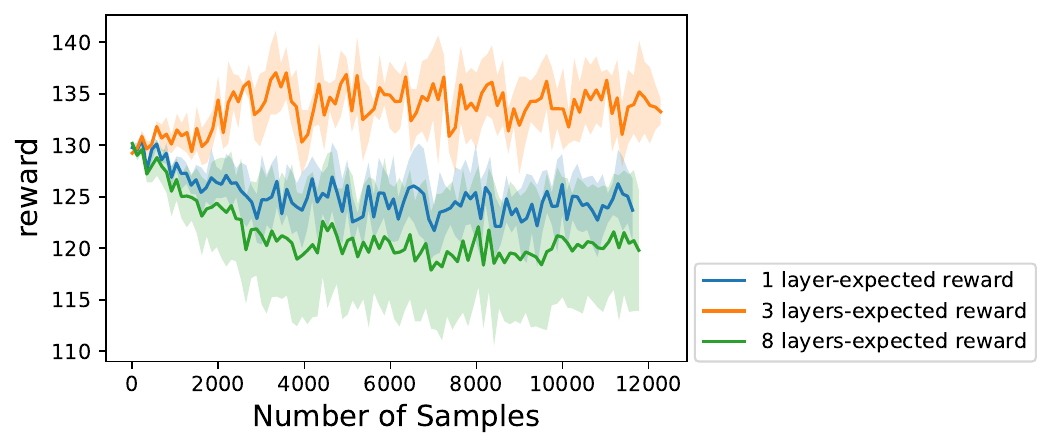}
	\hfill \caption{Role of data size in DQN with expected reward}
	\label{samples}
\end{figure}

Finally, we examine the role of sample size in DQN-e. Here we discard the used samples in the way described in the Beer Game experiment. We try DQN-e with 1 layer, 3 layers, and 8 layers. The results are reported in Figure \ref{samples}. It's clear that the performance of DQN-e with 3 layers improves to a stable point after the first 3000 samples, while DQN-e with 1 layer and 8 layers can't converge even after more than 10000 samples. 

\section{Conclusion}\label{Sec.Conclusion}
In this paper, we demonstrate the power of depth in deep Q-learning by showing its better generalization error bounds compared with those of the traditional version. Our main tools are a novel oracle inequality for Q-learning to show the importance of hypothesis spaces, two novel approximation theorems to show the expressive power of deep nets, and two generalization error estimates to exhibit the power of depth in deep Q-learning. We find that the main reason for the success of deep Q-learning is the outperformance of deep nets in approximating spatially sparse and piecewise smooth (or piecewise constant) functions, rather than
  its large capacity.  Our study has provided answers to Questions 1-3 in Sec. 1.1.

$\diamond$ {\bf Answer to Question 1.} As shown in Section 3.1, the most widely used reward functions in Q-learning are spatially sparse and piecewise constant (or piecewise smooth). Deep nets succeed in capturing these properties  (see Theorems \ref{Theorem:approx-piece-constant} and \ref{Theorem:piecewise-smooth}), which is beyond the capability of shallow nets or linear models \cite{chui1994neural,petersen2018optimal}. Thus, deep Q-learning performs much better than shallow nets and linear models in practice.

$\diamond$ {\bf Answer to Question 2.} As discussed in Sec. 3.2, deep Q-learning does not always outperform the traditional Q-learning. However, if reward functions in Q-learning possess certain sophisticated properties such as spatial sparseness, piecewise smoothness, piecewise constancy, and the properties in Table \ref{Tab:ReLU_fea},  then deep Q-learning performs better than shallow nets.

$\diamond$ {\bf Answer to Question 3.} The required sample size to finish a specific sequential decision-making problem depends on the properties of reward functions and the horizon $T$. Our results in Theorem \ref{Theorem:learning-piece-constant}, Theorem \ref{Theorem:learning-piece-smooth}, Corollary \ref{Corollary:learning-piece-constant}, and Corollary \ref{Corollary:learning-piece-smooth} quantified this relationship in terms of generalization error bounds.


\ifCLASSOPTIONcaptionsoff
  \newpage
\fi


\bibliographystyle{IEEEtran}
\bibliography{TPAMI}


%


%
%
%
%

\ifCLASSOPTIONcaptionsoff
  \newpage
\fi

\end{document}